%% file: main.tex
\documentclass[preprint,12pt,authoryear]{elsarticle}
\usepackage[utf8]{inputenc}
\usepackage[T1]{fontenc}
\usepackage{graphicx}
\usepackage{amsmath,amssymb,amsfonts}
\usepackage{booktabs}
\usepackage{pifont}
\newcommand{\cmark}{\ding{51}}
\newcommand{\xmark}{\ding{55}}
\usepackage{rotating}
\usepackage{makecell}
\usepackage{nicefrac}
\usepackage{dsfont}
\usepackage{microtype}
\usepackage{xcolor}
\usepackage{verbatim}
\usepackage{multirow}
\usepackage{tabularx}
\usepackage{makecell}
\usepackage{subcaption}
\usepackage{url}
\usepackage{hyperref}
\usepackage{float}
\usepackage{adjustbox}
\usepackage{array}

\journal{ISPRS Journal of Photogrammetry and Remote Sensing}
\begin{document}
\begin{frontmatter}
\title{DailyWildfireCAN: A Dataset and Benchmark for Next-Day Wildfire Spread Prediction in Canada}

\author[aff1,aff2]{Mahacine Ettahri\corref{cor1}\fnref{equal}}
\ead{mahacine.ettahri@etud.polymtl.ca}

\author[aff1]{Hager Radi Abdelwahed\fnref{equal}}

\author[aff1]{\\Toluwaleke Olutayo}

\author[aff2]{Daniel Aloise}

\author[aff1]{Gaétan Marceau Caron}
\affiliation[aff1]{
    organization={Mila - Quebec AI Institute},
    addressline={6666 Rue Saint-Urbain},
    city={Montreal, QC},
    postcode={H2S 3H1},
    country={Canada}
}

\affiliation[aff2]{
    organization={Computer and Software Engineering Department, Polytechnique Montreal},
    addressline={2500 Chemin de Polytechnique},
    city={Montreal,QC},
    postcode={H3T 1J4},
    country={Canada}
}
\cortext[cor1]{Corresponding author}
\fntext[equal]{Equal Contribution.}
\begin{abstract}
Wildfires represent a critical ecological and socio-economic challenge in Canada, where vast boreal forests and a warming climate have amplified the frequency and severity of extreme fire seasons.
Predicting next-day wildfire spread can support fire management, but requires capturing complex interactions among weather, fuels, and terrain. 
Deep learning has emerged as a promising approach to learning these relationships from heterogeneous environmental data, yet its development and evaluation depend on representative datasets. 
Existing deep learning research on wildfire spread relies predominantly on US-focused datasets, leaving Canadian fire environments underrepresented. 
To address this gap, we present \textbf{DailyWildfireCAN}, a multi-modal, multi-temporal dataset and benchmark for predicting next-day wildfire spread across Canada.
The dataset tracks continuous fire progression over multi-day events at a high spatial and temporal resolution of $90\text{ m}$ and $24\text{ h}$, respectively. 
This resolution preserves fine-scale information relevant to fire-edge geometry, local fuel variations, and terrain effects, supporting research toward improved daily wildfire spread forecasting. 
DailyWildfireCAN covers $2{,}440$ wildfire events across eight fire seasons ($2017\text{--}2024$) and integrates $33$ features across diverse modalities, including topography, meteorology, land cover, vegetation and forest structure, human influence indices, and satellite imagery. 
We establish a comprehensive benchmark comparing mono-temporal and multi-temporal prediction models, accompanied by a multi-dimensional evaluation framework assessing robustness across wildfire characteristics, geographic regions, and temporal conditions, spatial prediction quality, and sensitivity to reference label variations. 
Together, the dataset and evaluation framework provide a standardized foundation for identifying model limitations and advancing wildfire spread prediction in Canadian landscapes. We publicly release the dataset including raw data, mono-temporal and multi-temporal formats, along with benchmark models, trained checkpoints, and comprehensive evaluation code.

\end{abstract}

\end{frontmatter}

\section{Introduction}
\input{sections/intro}

\section{Related Work} \label{sec:related_work}
\input{sections/related_work}

\section{DailyWildfireCAN Dataset}
\input{sections/dataset}

\section{DailyWildfireCAN Benchmark}
\input{sections/benchmark}

\section{Evaluation Results and Analysis}
\input{sections/results}

\section{Conclusion}
\input{sections/conclusion}

\section*{Author Contributions: CRediT roles}

\noindent
\textbf{Mahacine Ettahri:}
Conceptualization, Methodology, Software, Validation,
Formal analysis, Investigation, Data curation, Resources, Writing -- original draft, Writing -- review \& editing, Visualization. \textbf{Hager Radi Abdelwahed:}
Conceptualization, Methodology, Software, Validation,
Formal analysis, Investigation, Resources, Writing -- original draft, Writing -- review \& editing, Visualization, Supervision, Project administration. \textbf{Toluwaleke Olutayo:}
Methodology, Software, Validation, Visualization, Writing -- review \& editing. \textbf{Daniel Aloise:}
Writing -- review \& editing, Supervision, Funding acquisition. \textbf{Gaétan Marceau Caron:}
Conceptualization, Methodology, Resources, Writing -- review \& editing, Supervision, Funding acquisition.

\section*{Acknowledgments}
The authors would like to thank Chirag Goel for his contributions to the early development of the research methodology, and Francisco Emiliano Lopez for his assistance in identifying and correcting issues in the forest disturbance maps. We also thank the researchers and staff at Natural Resources Canada (NRCan) who provided valuable clarifications and technical guidance regarding the Canadian Fire Spread Dataset (CFSDS) and the SCANFI forest inventory data. 

\section*{Funding}
This work was supported by the Institute for Data Valorization (IVADO), grant 2025-682883. The funding organizations had no role in the study design, data collection, analysis, interpretation of results, manuscript preparation, or decision to submit the article for publication. In addition, this research benefited from computing resources, software, and technical support provided by Mila (mila.quebec) and the Digital Research Alliance of Canada. 

\section*{Declaration of competing interest}
The authors declare that they have no known competing financial interests or personal relationships that could have appeared to influence the work reported in this paper.

\section*{Data availability}
The DailyWildfireCAN dataset is publicly available on
\href{https://huggingface.co/datasets/DailyWildfireCAN/DailyWildfireCAN}{Hugging Face},
along with the corresponding benchmark implementations and evaluation code
on \href{https://github.com/milatechtransfer/DailyWildfireCAN}{GitHub}.
Trained checkpoints for all benchmark models are also publicly available on
\href{https://huggingface.co/DailyWildfireCAN/DailyWildfireCAN-Benchmark}{Hugging Face}.
The dataset is released under the Creative Commons
Attribution-NonCommercial-ShareAlike 4.0 International
(CC BY-NC-SA 4.0) license, while incorporated third-party
materials remain subject to their respective original licenses.

\section*{Declaration of generative AI and AI-assisted
technologies use}

During the preparation of this work, the authors used LLMs to assist with improving the clarity and readability of the manuscript. The authors reviewed and edited the generated content as needed and take full responsibility for the content of the publication.

\bibliographystyle{elsarticle-harv}
\bibliography{references}
\appendix
\counterwithin*{table}{section}
\counterwithin*{figure}{section}

\renewcommand{\thetable}{\Alph{section}.\arabic{table}}
\renewcommand{\thefigure}{\Alph{section}.\arabic{figure}}
\input{sections/appendix.tex}

\end{document}

%% file: sections/intro.tex
Wildfires pose a major ecological and socio-economic challenge, particularly in Canada, where extensive boreal forests, remote communities, and complex fuel--weather interactions make wildfire behavior difficult to predict. The record-breaking 2023 wildfire season, during which approximately 15 million hectares burned, an area larger than England and more than double the previous national record of 6.7 million ha in 1989, demonstrated the scale of this challenge \citep{kirchmeier-young2024human, nrcan2023wildfires}. 
This was followed by the second-worst wildfire season on record in 2025, with approximately 8.9 million hectares burned \citep{eccc2025wildfires}, underscoring the persistent and growing need for improved wildfire monitoring and prediction. 
Therefore, reliable short-term predictions of wildfire spread are increasingly important for anticipating threats to communities and infrastructure, supporting timely evacuation decisions and prioritizing resource allocation, by forecasting how active fires may evolve over the following day(s).

Recent advances in machine learning have enabled wildfire growth and hazard prediction using large-scale geospatial observations, including satellite imagery, weather, vegetation, and terrain information \citep{huot2022next,gerard2023wildfirespreadts,zhao2025tssatfire,abdelwahed2026accelerating}. 
Notable datasets such as Next Day Wildfire Spread \citep{huot2022next} and WildfireSpreadTS \citep{gerard2023wildfirespreadts} have established benchmarks for short-term wildfire spread prediction. However, these datasets focus on U.S. wildfire events and operate at coarser spatial resolutions of 1,000 m and 375 m, respectively. Canadian datasets address related but different objectives: CanadaFireSat \citep{porta2025canadafiresat} focuses on wildfire occurrence forecasting through patch classification, while CFSDS \citep{CFSDS} provides reconstructed daily fire progression maps and associated environmental covariates, rather than a standardized benchmark for next-day wildfire spread prediction.
Despite these contributions, existing datasets do not jointly provide the spatially explicit fire-growth labels, consistent input features, and predefined data splits needed to support reproducible research on high-resolution next-day wildfire spread prediction across Canada.

Furthermore, comprehensive benchmarking of wildfire spread prediction models remains limited in the Canadian context. While existing studies have explored different architectures and evaluation strategies, comparisons spanning mono-temporal and multi-temporal models across diverse architectural families remain scarce. 
Similarly, systematic evaluations of model performance across geographic regions, temporal conditions, fire characteristics, and spatial prediction quality are limited. 
Such evaluations are important for understanding model capabilities and limitations beyond aggregate predictive performance, to inform their suitability for real-world applications and operational decision-making.

\begin{figure*}
  \centering
  \includegraphics[width=\textwidth]{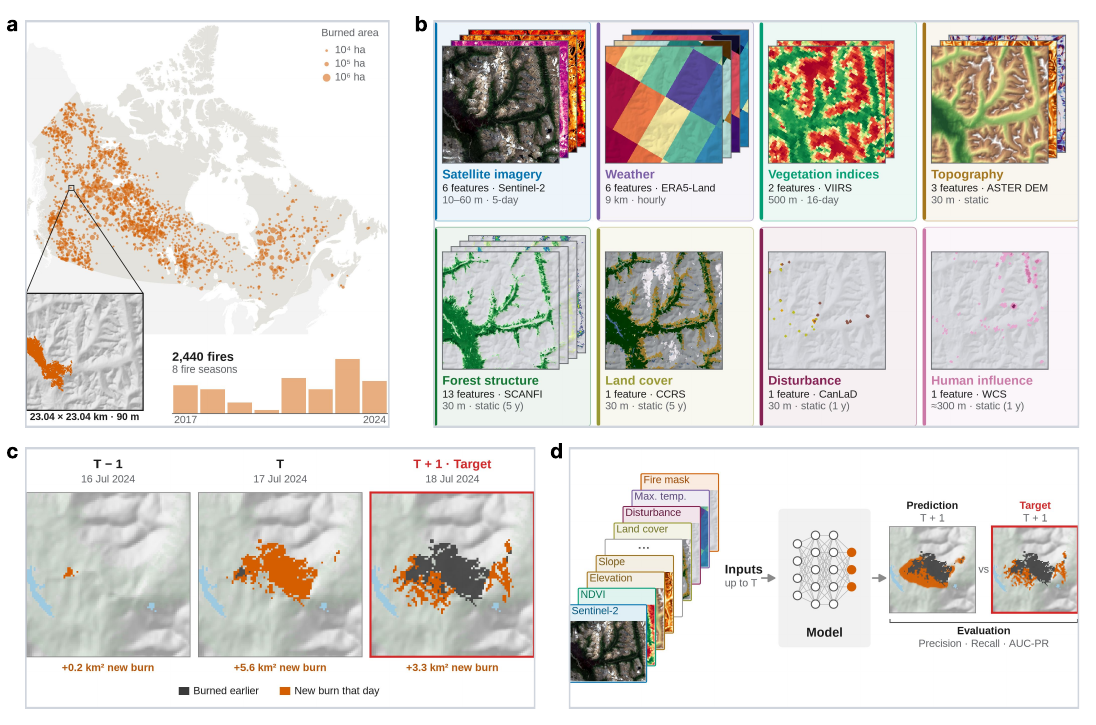}
    \caption{\textbf{DailyWildfireCAN Dataset and Benchmark.}
\textbf{(a) Geographic coverage.} Locations of the $2{,}440$ wildfires
(2017--2024), with marker size indicating total burned area.
Bars show the number of fires per year. The inset shows one data sample with its
burned area within a $23.04 \times 23.04$\,km
tile at $90$-m spatial grid.
\textbf{(b) Input features.} 
The $33$ input features for the example
tile, grouped by data source. 
Each group indicates the number of
features, source dataset, native spatial resolution and temporal
frequency.
\textbf{(c) Daily Fire Progression.} Evolution of an example fire
from 16 to 18 July 2024. Dark grey indicates previously burned
areas, while orange indicates newly burned areas on each day.
The newly burned area on day $T+1$ constitutes the prediction
target.
\textbf{(d) Benchmark.} Input features available up to
day $T$ are used to predict the newly burned area on day $T+1$.
Predictions are evaluated against the observed target using
precision, recall and area under the precision--recall curve
(AUC-PR). }
\label{fig:dataset_overview}
\end{figure*}

To address these gaps, we introduce \textbf{DailyWildfireCAN}, a comprehensive dataset and benchmark that formulates next-day wildfire spread prediction across Canada, using a standardized 90-m spatial grid. Figure~\ref{fig:dataset_overview} provides an overview of DailyWildfireCAN, illustrating the prediction task and dataset structure. Our contributions are summarized as follows:

\begin{itemize}
    \item We formulate next-day wildfire spread prediction for Canadian wildfires as a pixel-wise segmentation task, targeting the spatial growth of ongoing fires over the following 24 hours.
    \item We construct a dataset comprising $2,440$ wildfire events across eight Canadian fire seasons (2017--2024), organized into $23.04$ km $\times$ $23.04$ km spatial patches at $90$ m pixel size, integrating daily fire progression with multimodal environmental and satellite observations in both mono-temporal and multi-temporal formats.
    \item We establish a reproducible benchmark comparing diverse mono-temporal and multi-temporal model architectures, providing predefined data splits, baseline implementations, trained checkpoints, and consistent evaluation protocols.
    \item We provide a multi-dimensional evaluation of model generalization and robustness across fire characteristics, geographic regions, temporal conditions, spatial prediction tolerances, and label variations, revealing limitations beyond aggregate predictive performance.
\end{itemize}

We expect DailyWildfireCAN to support the development of more robust and generalizable wildfire spread prediction models and advance research at the intersection of machine learning and environmental science. 
The remainder of this paper reviews related wildfire datasets and benchmarks, details the construction of DailyWildfireCAN and its standardized evaluation framework, presents benchmark results and analyses of model robustness, generalization, and spatial prediction quality, and concludes with key limitations and directions for future research.

%% file: sections/related_work.tex
\newcommand{\tmark}{\ensuremath{\circledcirc}}
\begin{table}[!htbp]
\centering
\caption{Comparison of DailyWildfireCAN with related wildfire datasets and benchmarks, discussed in Section \ref{sec:related_work}, across geographic coverage, task, spatial grid size, satellite/spectral inputs, and 2D spatial environmental inputs: weather (\textit{W}), terrain (\textit{T}), vegetation/land cover (\textit{V}) - including explicit greenness, fuel, or land-cover products, forest structure (\textit{F}) - including stand metrics such as height, canopy closure, and biomass, forest disturbance history (\textit{D}) - referring to land disturbances such as defoliation, wildfires, or harvesting over the past year, and human influence (\textit{H}).}
\label{tab:benchmark_comparison}
\begingroup
\fontsize{7.5}{9}\selectfont
\setlength{\tabcolsep}{1.4pt}
\renewcommand{\arraystretch}{1.04}
\hypersetup{hidelinks}

\begin{tabularx}{\linewidth}{@{}
  >{\raggedright\arraybackslash}p{26.5mm}
  >{\raggedright\arraybackslash}p{9.5mm}
  c
  r
  c
  >{\raggedright\arraybackslash}p{21.5mm}
  >{\raggedright\arraybackslash}X
  *{6}{>{\centering\arraybackslash}p{2.6mm}}@{}}
\toprule
& & & & & & & \multicolumn{6}{c}{\textbf{2D Env.\ inputs}} \\
\cmidrule(l){8-13}
\textbf{Dataset\textsuperscript{a}} &
\textbf{Region} &
\textbf{Period} &
\textbf{\# Fires} &
\textbf{Grid} &
\textbf{Task} &
\textbf{Spectral\newline inputs} &
\textbf{W} &
\textbf{T} &
\textbf{V} &
\textbf{F} &
\textbf{D} &
\textbf{H} \\
\midrule

Next Day Wildfire Spread
  & USA & 2012--2020 & --- & 1\,km
  & Active-fire prediction (1-day)
  & VIIRS index
  & \cmark & \cmark & \cmark & \xmark & \xmark & \cmark \\
\addlinespace[2pt]

WildfireSpreadTS
  & USA & 2018--2021 & 607 & 375\,m
  & Active-fire prediction (1-day)
  & VIIRS bands, indices
  & \cmark & \cmark & \cmark & \xmark & \xmark & \xmark \\
\addlinespace[2pt]

TS-SatFire (progression)
  & USA & 2017--2021 & 179 & 375\,m
  & Burned-area growth (1-day)
  & VIIRS bands, indices
  & \cmark & \cmark & \cmark & \xmark & \xmark & \xmark \\
\addlinespace[2pt]

CanadaFire2023
  & Canada & 2023 & --- & 10/30\,m\textsuperscript{b}
  & Burned-area mapping
  & Landsat-8/9, Sentinel-2 bands, indices
  & \xmark & \xmark & \xmark & \xmark & \xmark & \xmark \\
\addlinespace[2pt]

WildfireCube
  & USA, Canada, Greece & 2017--2023 & 13 & 30\,m
  & Fire progression (3-hour)
  & Sentinel-2 indices
  & \cmark & \cmark & \cmark & \xmark & \xmark & \xmark \\
\addlinespace[2pt]

BCWildfire
  & BC \& adjacent areas & 2000--2024 & --- & 1\,km
  & Fire-occur\-rence (1-day)
  & MODIS bands, indices
  & \cmark & \cmark & \cmark & \cmark & \xmark & \cmark \\
\addlinespace[2pt]

CFSDS\textsuperscript{c}
  & Canada & 2002--2024 & 5{,}435 & 90\,m
  & Fire-pro\-gres\-sion (1-day)
  & \xmark
  & \tmark & \tmark & \tmark & \tmark & \xmark & \tmark \\
\addlinespace[2pt]

CanadaFireSat
  & Canada & 2016--2023 & --- & 100\,m
  & Fire-occur\-rence (8-day)
  & Sentinel-2 bands; MODIS indices
  & \cmark & \xmark & \cmark & \xmark & \xmark & \xmark \\
\addlinespace[2pt]
\midrule

\textbf{DailyWildfireCAN}\newline \textbf{(ours)}
  & Canada & 2017--2024 & 2{,}440 & 90\,m
  & Fire Spread (1-day)
  & Sentinel-2 bands; VIIRS indices
  & \cmark & \cmark & \cmark & \cmark & \cmark & \cmark \\
\bottomrule
\end{tabularx}

\par\vspace{3pt}
\begin{minipage}{\linewidth}
\fontsize{7.5}{9}\selectfont
\raggedright
---: distinct-fire count not reported;\\
\cmark: spatially complete 2D raster input;
\tmark: georeferenced tabular covariates;
\xmark: not included; \\
\textsuperscript{a} All datasets include a benchmark, except for WildfireCube and CFSDS. \\
\textsuperscript{b}CanadaFire2023 provides segmentation datasets at 10\,m for Sentinel-2 and 30\,m for Landsat-8/9.\\
\textsuperscript{c}Originally published at 180\,m (2002--2021) and updated later to 90\,m (2002--2024) \\
BC: British Columbia, Canada. \\
\end{minipage}
\endgroup
\end{table}

In this section, we review the wildfire forecasting, spread prediction, and burned-area mapping datasets most closely related to DailyWildfireCAN. 
We organize the comparison around four aspects that are central to the design of a wildfire spread benchmark: a) geographic coverage, b) prediction target, c) spatial resolution, and d) input modality, as summarized in Table~\ref{tab:benchmark_comparison}.

\textbf{Next Day Wildfire Spread.} \citet{huot2022next} introduced the Next Day Wildfire Spread dataset, a public benchmark for day-ahead wildfire spread prediction over the United States.
In this dataset, the model is given the current active-fire mask together with environmental covariates and is asked to predict the active-fire mask one day later.
The fire labels are derived from the Moderate Resolution Imaging Spectroradiometer (MODIS), a satellite sensor aboard NASA's Terra and Aqua platforms, using the MODIS daily fire product \citep{giglio2015mod14a1}, and all variables are aligned to a $1$ km grid.
While this dataset established a standardized protocol for machine learning in wildfire spread prediction, the $1$ km grid limits its suitability for fine-scale fire-growth modeling.
At this scale, a single pixel aggregates substantial variation in fuels and terrain, reducing the benchmark's ability to represent local heterogeneity relevant to spread dynamics.
The authors acknowledge this limitation, noting that the MODIS-based representation does not provide the fine-scale information needed for tactical decision-making.
Thus, its U.S. coverage, active-fire prediction target, and $1$ km grid distinguish it from DailyWildfireCAN, which predicts next-day newly burned area across Canada on a $90$ m grid.

\textbf{WildfireSpreadTS.} \citet{gerard2023wildfirespreadts} proposed WildfireSpreadTS, which represents each fire event as a multi-day image time series over the United States, for a total of $607$ wildfire events.
The inputs include previous active-fire maps from the Visible Infrared Imaging Radiometer Suite (VIIRS), a satellite sensor used for active-fire detection \citep{viirs_active_fire}, together with VIIRS reflectance and vegetation products, weather variables, weather forecasts, topography, and land-cover information.
Relative to Next Day Wildfire Spread, WildfireSpreadTS provides a finer spatial grid of $375$ m.  
However, its prediction target remains a next-day VIIRS active-fire map. 
This distinction is important: an active-fire map records locations where the satellite detected active burning during an overpass, whereas a daily growth mask records the area newly burned between two consecutive days. 
Thus, although WildfireSpreadTS provides a richer temporal and multimodal formulation than earlier benchmarks, it differs from DailyWildfireCAN in its U.S. coverage, active-fire prediction target, and $375$ m grid.

\textbf{TS-SatFire.} \citet{zhao2025tssatfire} presented TS-SatFire, a multi-task satellite time-series benchmark for active-fire detection, burned-area mapping, and wildfire progression prediction. 
Its progression task is closely aligned with our target definition, as it predicts newly burned area between the last observed day and the next day. 
However, TS-SatFire remains centered on U.S. wildfire events and is built around VIIRS-based observations resampled to a $375$ m grid. 
Thus, while TS-SatFire addresses a similar next-day progression target, it differs from our benchmark in geographic scope, sensor basis, and spatial grid resolution.

\textbf{CanadaFire2023.} \citet{zhong2024canadafire2023} introduced CanadaFire2023, a benchmark for burned-area mapping from satellite imagery acquired across Canada between 1 May and 1 November 2023. 
It contains separate Landsat-8 \citep{roy2014landsat8} and Landsat-9 \citep{masek2020landsat9} subsets on $30$ m grids and a Sentinel-2 subset on a $10$ m grid, with manually annotated burn-scar masks. 
Inputs comprise multispectral imagery augmented with the Normalized Burn Ratio \citep{key2006nbr} and the Normalized Difference Water Index \citep{mcfeeters1996ndwi}.
Although its Canadian coverage and Sentinel-2 imagery are relevant to DailyWildfireCAN, its task is to segment areas already burned at the time of image acquisition, rather than to predict the next day's spatial growth of an ongoing fire.
Its temporal coverage is also restricted to the 2023 fire season, whereas DailyWildfireCAN spans eight fire seasons from 2017 to 2024.

\textbf{WildfireCube.} \citet{linardos2026wildfirecube} introduced WildfireCube, a pipeline for assembling time sequences of spatially aligned environmental and fire-progression maps. 
It combines terrain variables derived from the Copernicus global $30$ m elevation model (GLO-30) \citep{copernicusdem}, ERA5-Land reanalysis weather \citep{munoz2021era5}, vegetation and burn-severity indices derived from Sentinel-2 optical imagery \citep{drusch2012sentinel2}, and MODIS/VIIRS active-fire detections.
Fire-arrival times are reconstructed by spatial interpolation and grouped into $3$-hour intervals to produce fire-growth maps on a $30$ m grid. 
The study covers $13$ fires across the United States, Canada, and Greece during 2017--2023. 
Although WildfireCube provides a finer spatial grid and sub-daily progression, DailyWildfireCAN offers substantially greater Canadian event coverage, comprising $2{,}440$ fires on a $90$ m grid for next-day spatial growth prediction. DailyWildfireCAN also incorporates explicit forest-structure, land-cover, disturbance, and human-influence inputs.

\textbf{BCWildfire.} \citet{xu2026bcwildfire} introduced BCWildfire, a multi-factor dataset for wildfire risk prediction in British Columbia covering the period from 2000 to 2024. 
BCWildfire incorporates weather, topography, fuel dynamics, and human factors, and is primarily built on MODIS and VIIRS satellite sensors. 
However, its target is fire-occurrence risk, whereas DailyWildfireCAN predicts next-day spatial growth for an ongoing fire.
It is also geographically restricted to British Columbia and uses a $1$ km grid, compared with DailyWildfireCAN's national coverage and $90$ m grid.

\textbf{Canadian Fire Spread Dataset (CFSDS).} CFSDS \citep{CFSDS} is highly relevant to our dataset because it provides daily fire-progression information for Canadian wildfires. 
It combines the National Burned Area Composite (NBAC) \citep{skakun2022extending} fire perimeters with active-fire detections from MODIS and VIIRS satellite sensors to estimate fire-arrival times for large wildfire events. 
The original release covers fires larger than $1000$ ha in forested areas of Canada from 2002 to 2021 at $180$ m spatial resolution, while the latest release on the Open Science Framework (OSF) increases the spatial resolution to $90$ m, lowers the size threshold to $500$ ha, and includes fires through 2024. 
CFSDS provides georeferenced records of burned pixels, including their coordinates and estimated days of burning; its environmental covariates are supplied as tabular attributes associated with those burned pixels. This representation requires additional data processing and spatial alignment to construct training-ready samples for deep learning models that predict wildfire spread across the surrounding landscape. Furthermore, CFSDS does not provide predefined training, validation, and test splits or a standardized benchmark for next-day wildfire spread prediction. In contrast, DailyWildfireCAN provides spatially aligned input maps covering both active fires and their surrounding landscapes, paired with next-day growth targets, predefined data splits, a benchmark and reproducible evaluation protocols.

\textbf{CanadaFireSat.} CanadaFireSat \citep{porta2025canadafiresat} is a closely related Canadian dataset and benchmark. 
It uses Sentinel-2 optical image time series, together with coarser environmental and fire-weather predictors from MODIS, ERA5-Land data, and Copernicus Emergency Management Service (CEMS) fire-danger products. 
Labels are derived from NBAC \citep{skakun2022extending}, and examples are defined over $2.64$ km $\times$ $2.64$ km regions at $100$ m resolution. 
The key difference is the prediction target: CanadaFireSat predicts whether fire occurs anywhere within a region during the next $8$ days, whereas DailyWildfireCAN predicts next-day spatial growth for an already ongoing fire.
The covariates also differ: CanadaFireSat does not include topographic variables such as elevation, slope, and aspect, which DailyWildfireCAN provides to represent terrain relevant to local spread dynamics.

\textbf{Positioning of DailyWildfireCAN.} 
As shown in Table~\ref{tab:benchmark_comparison}, the datasets discussed above provide complementary resources for wildfire forecasting and mapping, but leave a gap for fine-scale daily wildfire growth prediction across Canada.
The remaining need is a national benchmark that pairs next-day newly burned-area targets with spatially aligned inputs describing both an ongoing fire and the surrounding landscape.
DailyWildfireCAN addresses this need by combining Canadian wildfire events, Sentinel-2 imagery, and a comprehensive suite of rasterized environmental covariates, spanning weather, terrain, vegetation, forest structure, forest disturbance history, and human influence, on a $90$\,m spatial grid over $23.04\,\mathrm{km} \times 23.04\,\mathrm{km}$ regions.

%% file: sections/dataset.tex
\label{sec:dataset}
This section details the construction of DailyWildfireCAN, covering fire progression labels, input feature layers, spatial alignment, sample generation and filtering, dataset splits, and preprocessing procedures.

\subsection{Data Sources}
Table~\ref{tab:dataset_features} summarizes the input features and output labels, including their data sources, original spatial and temporal resolutions. Each feature group is described in detail below.

\subsubsection{Daily Wildfire Labels}
DailyWildfireCAN uses CFSDS \citep{CFSDS} as the source of daily fire-progression information. 
We select wildfire events from 2017 to 2024 with a final burned area greater than $500$\,ha, resulting in $2{,}440$ fires. 
For each event, CFSDS provides an estimated fire-arrival day for burned locations within the final fire perimeter. 
We rasterize these arrival times onto a common grid of height $H$ and width $W$ (in pixels) at a $90$\,m spatial resolution to construct daily binary fire masks.
Let \(C_t \in \{0,1\}^{H \times W}\) denote the cumulative burned-area mask on day \(t\), where pixels burned on or before day \(t\) are assigned a value of \(1\), and all other pixels in the fixed spatial region are assigned \(0\). 
In DailyWildfireCAN, we set \(H=W=256\) pixels, corresponding to a \(23.04\,\mathrm{km} \times 23.04\,\mathrm{km}\) region. 
The target for one-day-ahead spread prediction is the next-day fire-growth mask \(Y_{t+1} \in \{0,1\}^{H \times W}\), defined at pixel coordinates \((i,j)\) as:
\begin{equation}
\label{eq1}
Y_{t+1}(i,j) = \mathds{1}
\{C_t(i,j)=0,\; C_{t+1}(i,j)=1\},
\quad 1 \leq i \leq H,\; 1 \leq j \leq W,    
\end{equation}
where \(\mathds{1}\{\cdot\}\) denotes the indicator function. 
This mask isolates pixels newly burned between days \(t\) and \(t+1\), signaling active fire spread. We use the cumulative burned-area mask \(C_t\) as an input feature representing the current footprint of the fire.

\begin{table}[htbp]
\centering
\caption{Summary of input features and wildfire labels
used in DailyWildfireCAN, including their sources,
feature counts, native spatial resolutions, and
temporal frequencies. All inputs are
harmonized to a common 90 m spatial grid in DailyWildfireCAN.}
\label{tab:dataset_features}

\footnotesize
\setlength{\tabcolsep}{4pt}
\renewcommand{\arraystretch}{1.15}

\begin{tabularx}{\linewidth}{
    >{\raggedright\arraybackslash}X
    >{\raggedright\arraybackslash}X
    c c c
}
\toprule
\textbf{Feature Group} &
\textbf{Source} &
\textbf{Features} &
\makecell{\textbf{Native}\\\textbf{Resolution}} &
\makecell{\textbf{Temporal}\\\textbf{Frequency}} \\
\midrule

\multicolumn{5}{@{}l}{\textbf{Environmental Features}} \\
\addlinespace[2pt]

Topography (elevation, slope, aspect)
& ASTER DEM \citep{aster_dem}
& 3 & 30 m & Static \\

Weather (temperature, humidity,
precipitation, wind (u, v))
& ERA5-Land \citep{munoz2021era5}
& 6 & 9 km & Hourly \\

Vegetation indices (NDVI, EVI)
& VIIRS \citep{VIIRS_VNP13A1_2021}
& 2 & 500 m & 16-day \\

Forest structure and composition
& SCANFI \citep{scanfi}
& 13 & 30 m & Static (5 y) \\

Land cover
& CCRS \citep{landcover_ccrs}
& 1 & 30 m & Static (5 y)\\

Forest Disturbance
& CanLaD \citep{canlad}
& 1 & 30 m & Static (1 y) \\

Human Influence Index
& WCS \citep{sanderson2022march}
& 1 & $\sim$300 m & Static (1 y) \\

\midrule
\multicolumn{5}{@{}l}{\textbf{Satellite Imagery}} \\
\addlinespace[2pt]

Multispectral (RGB, NIR, SWIR)
& Sentinel-2A \citep{drusch2012sentinel2}
& 6 & 10--60 m & 5-day \\

\midrule
\multicolumn{5}{@{}l}{\textbf{Wildfire Labels}} \\
\addlinespace[2pt]

Daily fire progression mask
& CFSDS \citep{CFSDS}
& -- & 90 m & Daily \\

\bottomrule
\end{tabularx}
\end{table}

\subsubsection{Input Features: Environmental}
We describe the environmental input layers and their corresponding data preparation procedures below. 
All raster layers are reprojected to EPSG:3347 and resampled to a common \(90\) m spatial grid aligned with the fire-growth labels.

\textbf{Topography:} Topographic variables are derived from the NASA ASTER Global Digital Elevation Model V3, a \(30\) m digital elevation model (DEM) \citep{aster_dem}.
We download DEM tiles covering the Canadian bounding box, reproject and downsample them to our target \(90\) m resolution using average resampling.
We extract elevation data and compute slope and aspect from the resulting elevation grids using the Horn algorithm \citep{horn}. Together, these topographic features capture how terrain modulates upslope heat transfer, solar exposure, and local fuel-moisture conditions.
Although local terrain can change due to rare disturbances or data updates, elevation, slope, and aspect are effectively static at the spatial and temporal scales considered here. 
Therefore, we treat all three as time-invariant inputs across the 2017--2024 period.

\textbf{Weather:} Meteorological variables are obtained from ERA5-Land reanalysis data \citep{munoz2021era5}, which provides hourly land-surface and near-surface atmospheric fields at approximately \(9\) km resolution.
Following the methodology of \citet{CFSDS}, we aggregate the raw hourly records into six daily metrics: maximum temperature over the 24-hour period, accumulated precipitation from noon of the previous day to noon of the current day, relative humidity at noon, noon wind speed at 10-m elevation, and the eastward and northward components of 10-m wind velocity. 
To align with our target \(90\) m grid, we spatially resample the ERA5-Land data using nearest-neighbor interpolation. 
These variables capture short-term atmospheric controls on intensity of fire, spread rate, and spread direction.

\textbf{Vegetation Indices:} We characterize vegetation conditions using the NASA VIIRS 16-Day vegetation indices product \citep{VIIRS_VNP13A1_2021}, available at \(500\) m resolution with a 16-day compositing period.
We acquire Normalized Difference Vegetation Index (NDVI) and Enhanced Vegetation Index (EVI) via Google Earth Engine as indicators of vegetation greenness and condition, providing indirect information about live fuel availability and moisture.
For each daily sample, we use the most recent 16-day composite whose observation period ends before the prediction date, ensuring that no future information is introduced. 
We resample the vegetation indices to \(90\) m using nearest-neighbor interpolation.

\textbf{Forest Structure and Composition:} We complement the VIIRS vegetation indices with forest-structure variables from Spatialized CAnadian National Forest Inventory (SCANFI v2) \citep{scanfi,guindon2026scanfi}, a product of Natural Resources Canada, available at \(30\) m resolution and provided in five-year increments.
Unlike NDVI and EVI, which provide spectral indicators of vegetation condition, SCANFI provides inventory-derived information about the underlying forest structure and composition.
We incorporate aboveground tree biomass, forest height, and total crown closure to characterize forest structure. Aboveground biomass, measured in tonnes per hectare, is derived from provincial merchantable-volume estimates. Forest height, measured in meters, characterizes the vertical extent of the tree canopy, while crown closure represents the percentage of each pixel covered by tree canopy.
To capture species composition, we include canopy cover fractions such as the percentage of broadleaf tree cover (\(\texttt{prcB}\)), the percentage of balsam fir crown (\(\texttt{prc\_balsam\_fir}\)), and the percentage of coniferous tree cover (\(\texttt{prcC}\)).
Together, these variables capture fuel availability, canopy structure, and forest composition, which influence fire behavior, including potential spread rate and intensity.
Since SCANFI maps are updated every five years, we use the 2015 products for 2017--2020 fires and the 2020 products for 2021--2024 fires. 
The native \(30\) m rasters are downsampled to \(90\) m using average resampling.

\textbf{Land Cover:} To complement SCANFI, we incorporate land cover classes from the Canada Centre for Remote Sensing \citep{landcover_ccrs}.
These classes provide information about land cover types and fuel continuity, which influence fire behavior, including potential spread rate and intensity. 
Specifically, the 15 land cover classes include needleleaf, broadleaf deciduous, and mixed forests; shrublands; grasslands; lichen-moss vegetation; wetlands; croplands; barren lands; urban and built-up areas; water; and snow and ice. 
We apply the same five-year temporal selection strategy as SCANFI to align land cover maps with each fire event.
The native \(30\) m maps are downsampled to \(90\) m using mode resampling to preserve their categorical nature.

\textbf{Forest Disturbance Maps:} To incorporate annual changes in forest conditions, we integrate CanLaD disturbance maps \citep{canlad}, available at \(30\)-m resolution, which distinguish five main disturbance categories: wildfires, harvesting, water extension, defoliation, and other disturbances (such as windthrow, partial harvesting, landslide, and unconfirmed wildfires)\footnote{We identified potential data leakage in the original disturbance maps, which are generated retrospectively and may reveal the future burned extent of active wildfires. To correct this leakage, we reconstructed the wildfire disturbance layer using only NBAC fire polygons \citep{skakun2022extending} from years preceding the prediction year, excluding the ongoing fire season. Only this corrected wildfire disturbance category is used for benchmark training, while the raw dataset preserves all original disturbance categories for future research and correction.}.
For each fire event, we construct historical wildfire disturbance layers using NBAC \citep{skakun2022extending} fire polygons from years preceding the prediction year, accumulating disturbances back to the reference year of the corresponding SCANFI and land cover products. This prevents information leakage from retrospectively generated disturbance maps. The original CanLaD disturbance categories are preserved in the raw dataset for future research.

\textbf{Anthropogenic Factors (Human Influence Index):} To quantify human pressure, we include the Human Influence Index (HII) \citep{sanderson2022march} as an input feature. 
The HII aggregates eight measures of human presence, including infrastructure, population density, and land use, into a single composite score ranging from 0 (no influence) to 64 (maximum influence) \citep{hii_site}. 
The HII maps are updated annually, allowing us to account for year-to-year changes in human influence. 
We resample the native \(\sim300\) m maps to \(90\) m using nearest-neighbor interpolation.

\subsubsection{Input Features: Satellite Imagery}
We use Sentinel-2 Level-2A (L2A) optical imagery, acquired via Microsoft Planetary Computer, to capture fine-scale surface conditions relevant to wildfire spread.
Sentinel-2 provides multispectral observations at \(10\)--\(60\) m resolution with a nominal five-day revisit cycle, although usable acquisitions over Canada depend on orbital overlap, acquisition geometry, and cloud cover.

We include six spectral bands: Blue (B02), Green (B03), Red (B04), Near-Infrared (NIR; B08), and two Short-Wave Infrared bands (SWIR; B11 and B12).
The visible bands capture surface reflectance patterns and burn-related changes, while NIR provides information about vegetation structure, greenness, and fire-affected areas.
The SWIR bands are sensitive to vegetation water content, fuel dryness, burn severity, and charred surfaces, making them particularly useful for wildfire growth modeling.
Because Sentinel-2 lacks daily temporal resolution, we use imagery acquired on day \(t\), when available, as input for predicting wildfire spread at \(t+1\). 
Otherwise, we select the most recent available acquisition within the preceding 29 days.
Acquisitions are considered chronologically, starting from the target date and moving backwards, ensuring that no observations acquired after the target date are used and preventing future-information leakage.
For each image, we preserve the acquisition date and cloud cover percentage in the raw files to support temporal alignment and flexible data filtering.
The multispectral bands are resampled using nearest-neighbor interpolation to match our common \(90\) m grid, producing spatially aligned \(256 \times 256\) image patches.
These features complement the environmental inputs by providing fine-scale observations of vegetation, fuel conditions, and landscape changes.

\subsection{Sample Construction and Filtering}
To construct DailyWildfireCAN, we partition the Canadian land area into non-overlapping \(256 \times 256\) spatial windows on the \(90\) m grid, each covering approximately \(23.04\,\mathrm{km} \times 23.04\,\mathrm{km}\). 
We retain only windows containing active wildfire events, as our benchmark focuses on predicting next-day fire growth from an already active fire. Wildfire ignition prediction is therefore outside the scope of DailyWildfireCAN.

Before defining the train, validation, and test splits, we apply two filtering steps: a) we remove one-day fire events, as they lack consecutive daily observations needed to construct next-day growth targets, b) we exclude samples with missing values in any required input layer. 
To support future research on missing-data handling, our released raw dataset includes a dedicated binary quality grid, where \(1\) indicates a missing feature value and \(0\) indicates valid data.
After filtering\footnote{To facilitate further research, we release both the unfiltered, unnormalized raw data and the final filtered samples after splitting.}, the dataset contains \(2,274\) active wildfire events.

\subsection{Data Preprocessing}
To prepare the dataset for benchmark model training, we preprocess and normalize input features using statistics computed exclusively from the training set. 
Continuous environmental variables, including weather, forest-structure features, elevation, and slope, are standardized using z-score normalization, while SCANFI percentage features are scaled to $[0,1]$ using min-max normalization. The Human Influence Index (HII) is transformed using \(\log(1 + x)\) to reduce the skewness of its distribution.
Terrain aspect is encoded using sine and cosine components to preserve its circular nature, replacing the original angle with two channels. 
NDVI and EVI are retained in their original scale, as their values are approximately bounded within $[-1,1]$. 
Sentinel-2 optical bands are scaled by $10^{-4}$ to convert stored integer values to surface reflectance, following the standard Sentinel-2 reflectance convention \citep{drusch2012sentinel2}. 
Finally, categorical land cover variables are one-hot encoded into $15$ binary channels, while wildfire disturbance is represented as a single binary channel indicating the presence (1) or absence (0) of historical wildfire disturbance.
We also include the current fire state as part of the model input using two auxiliary channels. 
The first is the cumulative burned-area mask \(C_t\) from \eqref{eq1}, which indicates pixels burned on or before day \(t\).
The second channel is a normalized fire-arrival-time mask, which encodes when each already burned pixel first burned relative to the start of the fire. 
Specifically, let \(\tau(i,j)\) denote the number of days from the start of the fire to the day on which pixel \((i,j)\) first burned. 
For prediction day \(t\), this mask is assigned value \(\tau(i,j)/100\) if \(\tau(i,j)\leq t\), and \(0\) otherwise. 
Thus, among pixels already burned by day \(t\), larger values correspond to later fire-arrival times, while unburned pixels are assigned \(0\). 
The 100-day scaling factor serves as a practical upper bound for most fires.
Combining the environmental, Sentinel-2, and auxiliary fire-state channels yields a total of \(50\) input channels per sample, represented as a \(256 \times 256 \times 50\) raster tensor.

\begin{figure*}
    \centering
    \includegraphics[width=0.49\linewidth]{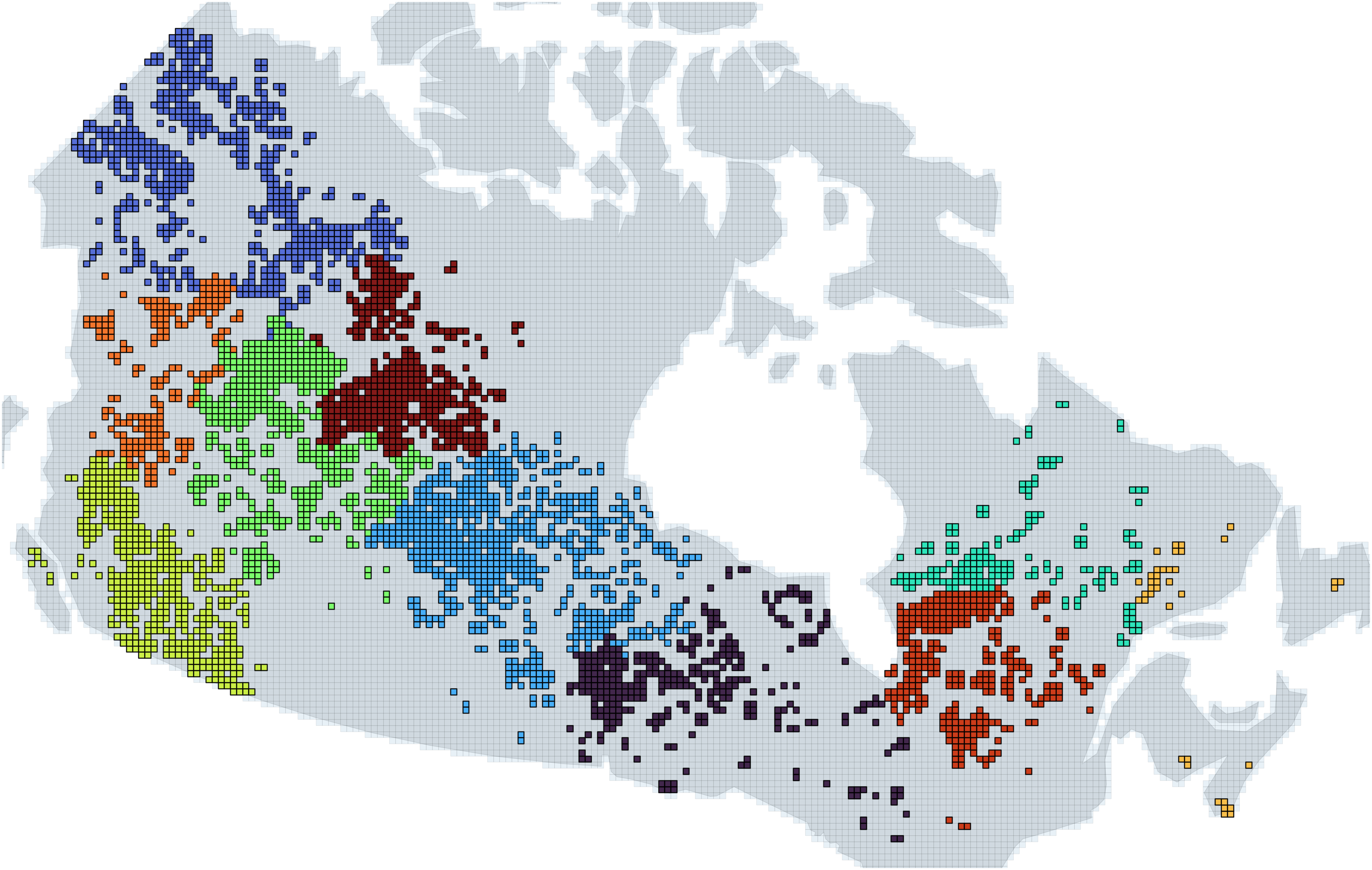}
    \includegraphics[width=0.49\linewidth]{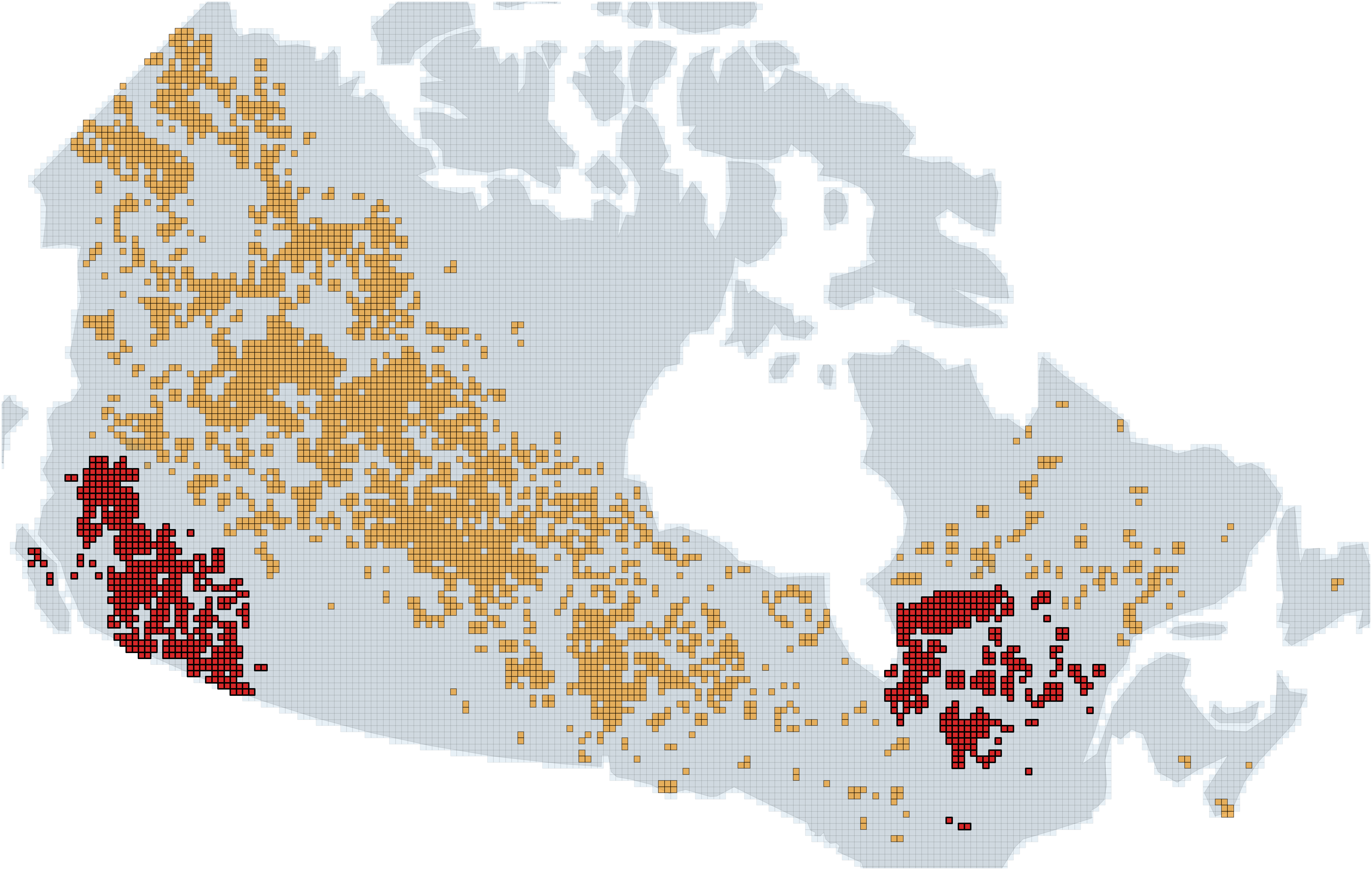}
    \caption{Spatial distribution of DailyWildfireCAN events: (a) spatial clusters across Canada and (b) training clusters (yellow) and held-out test clusters (red).}
    \label{fig:dataset_map}
\end{figure*}

\subsection{Dataset Splits} \label{sec:dataset_splits}
To evaluate the generalization capabilities of wildfire spread prediction models, we construct training, validation, and test splits across the 2017--2024 fire seasons using both spatial and temporal holdouts. 
We first group fire events into 10 spatial clusters using complete-linkage hierarchical agglomerative clustering \citep{johnson1967hierarchical}. 
We reserve 2024 as an unseen test year and select two spatial clusters, one from Western Canada and one from Eastern Canada, as held-out test regions, as shown in Figure~\ref{fig:dataset_map}. 
These clusters were chosen to capture distinct historical fire regimes, characterized by differences in fire size and duration. 
To reduce spatial autocorrelation and potential information leakage, we exclude all fire events within 50 km of the held-out clusters from the training and validation sets.

For the remaining eight spatial clusters during 2017--2023, we adopt a year-dependent spatial partitioning strategy, where one cluster is designated for validation in each year and the remaining seven clusters are used for training.
This design enables evaluation under three levels of generalization difficulty:

\begin{itemize}
    \item \textbf{Test-Easy (unseen year):} this includes fire events occurring in 2024 within regions represented during training, to evaluate temporal generalization to an unseen fire season.
    \item \textbf{Test-Medium (unseen regions):} this includes fire events occurring during 2017--2023 within the held-out spatial clusters, to evaluate spatial generalization to previously unseen geographical regions.
    \item \textbf{Test-Hard (unseen regions and year):} this includes fire events occurring in 2024 within the held-out spatial clusters, to evaluate simultaneous spatial and temporal generalization.
\end{itemize}

\subsection{Dataset Formats: Mono-Temporal and Multi-Temporal}
To accommodate different modeling approaches and facilitate research on wildfire propagation dynamics, we provide two complementary configurations of DailyWildfireCAN: mono-temporal and multi-temporal.

The \textit{mono-temporal version} serves as the primary benchmark, where each sample consists of observations from a single day used to predict the next-day wildfire spread mask. This formulation supports a broad range of conventional machine learning and deep learning architectures without requiring explicit temporal sequence modeling. The resulting dataset contains 81,271 samples spanning 2,274 fire events, with split statistics presented in Table~\ref{tab:dataset_stats_mono_temporal}. The resulting splits exhibit comparable distributions of fire size and duration, as further illustrated in ~\ref{app:dataset_analysis}.

\begin{table}[htbp]
\centering
\caption{Dataset statistics for DailyWildfireCAN: Mono-Temporal.}
\label{tab:dataset_stats_mono_temporal}
\begin{adjustbox}{width=\linewidth}
\begin{tabular}{lccccc}
\toprule
\textbf{Split} & \textbf{\# Samples} & \textbf{\# Fires} & \textbf{Time Span} & \textbf{Spatial Coverage} & \textbf{\%}\\
\midrule
Train & $46,123$ & $1,423$ & 2017--2023 & 7 clusters & $56.75\%$\\
\midrule
Validation & $3,760$ & $108$ & 2017--2023 & 1 cluster & $4.63\%$\\
\midrule
(1) Test-Easy & $16,052$ & $304$ & 2024 & 8 clusters & $19.75\%$\\
(2) Test-Medium & $13,911$ & $383$ & 2017--2023 & 2 clusters & $17.12\%$\\
(3) Test-Hard & $1,425$ & $56$ & 2024 & 2 clusters & $1.75\%$\\
\midrule
\textbf{Total} & $81,271$ & $2,274$ & 2017--2024 & 10 clusters & $100\%$ \\
\bottomrule
\end{tabular}
\end{adjustbox}
\end{table}
However, wildfire propagation is inherently dynamic, and a single snapshot of environmental conditions and the current fire extent may not fully capture the underlying physical state. 
In particular, recent precipitation and relative humidity influence fuel moisture and drying, while recent fire progression provides valuable information about spread direction, rate, and residual burning activity. These temporal dependencies may not be adequately represented by observations at a single time step. 
To investigate the benefits of incorporating historical context, we additionally provide a \textit{multi-temporal version}, where each sample consists of three consecutive days of environmental observations and fire progression masks used to predict the next-day wildfire spread. 
This configuration enables the evaluation of spatiotemporal models that explicitly leverage historical information to better characterize evolving fire dynamics.

Constructing three-day input sequences requires fire events with at least four consecutive days of observations, reducing the eligible dataset to $65,413$ samples from $2,063$ fire events, as summarized in Table~\ref{tab:dataset_stats_multi_temporal}. Both configurations preserve the same spatial and temporal splitting strategy, enabling evaluation under consistent generalization settings. Furthermore, we release the raw data, code and instructions to generate multi-temporal samples with different input sequence lengths ($n$ days), allowing the research community to investigate the effect of temporal context on wildfire spread prediction.

\begin{table*}[t]
\centering
\caption{Dataset statistics for DailyWildfireCAN: Multi-Temporal (3-day Timeseries).}
\label{tab:dataset_stats_multi_temporal}
\resizebox{\textwidth}{!}{
\begin{tabular}{lccccc}
\toprule
\textbf{Split} & \textbf{\# Samples} & \textbf{\# Fires} & \textbf{Time Span} & \textbf{Spatial Coverage} & \textbf{\%}\\
\midrule
Train & $36,696$ & $1,271$ & 2017--2023 & 7 clusters & $56.10\%$\\
\midrule
Validation & $3,107$ & $98$ & 2017--2023 & 1 cluster &  $4.75\%$\\
\midrule
(1) Test Easy & $13,242$ & $278$ & 2024 & 8 clusters & $20.24\%$\\
(2) Test Medium & $11,189$ & $364$ & 2017--2023 & 2 clusters & $17.11\%$\\
(3) Test Hard & $1,179$ & $52$ & 2024 & 2 clusters & $1.80\%$\\
\midrule
\textbf{Total} & $65,413$ & $2,063$ & 2017--2024 & 10 clusters & $100\%$ \\
\bottomrule
\end{tabular}
}
\end{table*}

%% file: sections/benchmark.tex
In this section, we describe the models used for training and evaluating DailyWildfireCAN, along with the evaluation metrics employed to assess their performance.

\subsection{Models}
To establish performance baselines, we evaluate a wide range of architectures commonly used for next-day wildfire spread prediction. 
We categorize these models into two groups: a) mono-temporal models, which rely on observations from a single time step, and b) multi-temporal models, which incorporate a short history of observations from preceding time steps.

\subsubsection{Mono-temporal Models.}
We evaluate a diverse set of approaches, ranging from simple heuristics and probabilistic fire-spread models to supervised deep learning architectures and pretrained foundation models. We organize these baselines into different categories: (1) heuristic and simulation-based baselines, (2) convolutional neural networks, (3) advanced neural architectures incorporating Transformers or state-space models, and (4) pretrained geospatial foundation models. This categorization enables a systematic comparison of different modeling approaches and architectural choices for wildfire spread prediction.

\paragraph{Heuristic and Simulation-based Baselines}
\begin{itemize}
    \item \textbf{Persistence Baseline}: We adopt a simple no-growth model as our primary non-ML baseline, assuming that the fire perimeter remains unchanged over the next day.
    \item \textbf{Dilation Baseline}: We introduce an additional non-ML baseline that expands the previous day's fire mask by one pixel in all directions, representing a simple uniform fire spread assumption.
    \item \textbf{Cellular Automata:} Following \citet{ALEXANDRIDIS2008191}, we implement a probabilistic cellular automata model that simulates cell-to-cell fire spread based on wind, slope, vegetation density, and non-burnable land cover. Similar cellular automata approaches are employed in wildfire simulators such as Cell2Fire \citep{cell2fire} and PROPAGATOR \citep{propagator}.
\end{itemize}

\paragraph{Convolutional Neural Networks}
\begin{itemize}
    \item \textbf{U-Net}: We use a standard U-Net \citep{ronneberger2015unet} as our main convolutional segmentation baseline, using an encoder--decoder architecture with skip connections.
    \item \textbf{U-Net with time-gap encoding}: We extend U-Net by adding a temporal-gap embedding representing the time difference between the satellite observation and prediction target, inspired by \citet{time_of_day_encoding}. This helps account for differences in acquisition times between satellite observations and prediction targets, particularly when only older imagery is available.
    \item \textbf{Attention U-Net}: We train Attention U-Net \citep{attention_unet}, which adds attention gates to U-Net skip connections to emphasize regions relevant to fire spread and suppress irrelevant background features.
\end{itemize}

\paragraph{Advanced Neural Architectures}
\begin{itemize}
    \item \textbf{SegFormer U-Net}: We include a hybrid U-Net--SegFormer baseline with a SegFormer-style hierarchical Transformer encoder \citep{seg_former}. This evaluates whether Transformer-based multi-scale features improve over a convolutional U-Net.
    \item \textbf{Swin-U-Net:} We incorporate Swin-U-Net \citep{swinunet2023}, a U-shaped encoder-decoder architecture with skip connections, where convolutional blocks are replaced with Swin Transformer blocks.
    \item \textbf{U-Mamba:} We include an encoder-decoder architecture that incorporates Mamba-based state-space blocks alongside traditional convolutions for feature extraction \citep{umamba2024}.
\end{itemize}

\paragraph{Geospatial Foundation Models}
We include two recent geospatial foundation models, \textbf{OlmoEarth-v1.2-Base} \citep{olmoearth2025} and \textbf{DINOv3} \citep{dinov3}, to assess the effectiveness of their learned representations for our next-day wildfire spread prediction task.

\subsubsection{Multi-temporal Models}
We evaluate a diverse set of multi-temporal architectures to investigate their ability to capture spatio-temporal dependencies in wildfire progression. All models use a three-day observation window to predict next-day wildfire spread.

\paragraph{Recurrent Architectures} 
\begin{itemize}
    \item \textbf{U-Net with ConvLSTM}: We integrate ConvLSTM \citep{convlstm} into a U-Net architecture to model temporal dependencies across a (3)-day sequence of observations, evaluating the benefit of recurrent spatio-temporal feature learning for next-day wildfire spread prediction.
\end{itemize}

\paragraph{Temporal Attention Architectures} 
\begin{itemize}
    \item \textbf{Unet with Temporal Attention Encoder (U-TAE)}: We include U-TAE \citep{UTAE}, which combines spatial convolutional features with temporal attention to aggregate information across (3)-day observation sequences.
    \item \textbf{Video Swin-U-Net:} We extend Swin-U-Net with Video Swin Transformer blocks \citep{videoswinunet2026} to capture spatial and temporal dependencies jointly across (3)-day observation sequences.
\end{itemize}

\paragraph{Convolutional Sequence Models}
\begin{itemize}
    \item \textbf{Simpler yet better video  prediction (SimVPv2)}: 
Originally designed for video prediction, SimVPv2 \citep{simvpv2} is adapted as a convolutional spatio-temporal forecasting baseline for next-day wildfire spread prediction.

\end{itemize}

\paragraph{Multi-Branch Fusion Architectures}
\begin{itemize}
    \item \textbf{A Multi-Modal Architecture:} Inspired by \citet{multimodal2024}, we include a CNN-based multi-branch architecture that separately processes daily weather variables, historical fire progression, and static geographical features such as topography and land cover. The extracted representations are combined through late fusion to predict next-day wildfire spread.
\end{itemize}

\subsection{Evaluation Metrics}
We evaluate our model's performance using standard binary segmentation metrics between the predicted and ground-truth wildfire masks. 
We include the following metrics: a) \textit{Precision}, to measure the fraction of predicted fire-spread pixels that are correct (with 0.5 threshold), penalizing false-positive predictions, b) \textit{Recall}, to measure the fraction of true fire-spread pixels detected by the model (with 0.5 threshold), capturing missed fire areas, and c) \textit{AUC-PR}: to measure the area under the precision--recall curve across prediction thresholds, making it useful for evaluating probabilistic fire-propagation predictions under class imbalance.

%% file: sections/results.tex
We present the results of the DailyWildfireCAN benchmark through a comprehensive evaluation of predictive performance, robustness and generalization, spatial prediction quality, and sensitivity to reference label variations. 
We first compare mono-temporal and multi-temporal models, followed by analyses of performance across wildfire characteristics, temporal conditions, and geographic regions. We further examine spatial accuracy at different boundary tolerances and spread distances, as well as the sensitivity of evaluation outcomes to spatial and temporal variations in reference fire progression labels. 
All models are trained under a consistent experimental setup to ensure fair comparison, with training details provided in ~\ref{appendix:training_details}. 
All models utilize the full set of available input features.

\newcommand{\meanstd}[2]{%
  \makecell[c]{{\normalsize $#1$}\\[-1pt]
  {\scriptsize $\pm #2$}}%
}

\newcommand{\bestmeanstd}[2]{%
  \makecell[c]{{\normalsize $\mathbf{#1}$}\\[-1pt]
  {\scriptsize $\pm #2$}}%
}

\begin{table}[htbp]
\centering
\caption{\textit{Mono-temporal} benchmark results for next-day wildfire spread prediction on DailyWildfireCAN. Performance metrics are reported as mean $\pm$ standard deviation across three random seeds. The best mean performance for each metric within each test set is highlighted in bold.}
\label{tab:benchmark_results_mono_temporal}

\small
\setlength{\tabcolsep}{2pt}
\renewcommand{\arraystretch}{1.1}

\begin{adjustbox}{width=\linewidth}
\begin{tabular}{@{}lccc|ccc|ccc@{}}
\toprule

\textbf{Model} &
\multicolumn{3}{c}{\textbf{Test-Easy}} &
\multicolumn{3}{c}{\textbf{Test-Medium}} &
\multicolumn{3}{c}{\textbf{Test-Hard}} \\
\cmidrule(lr){2-4}
\cmidrule(lr){5-7}
\cmidrule(lr){8-10}

&
\textbf{AUC-PR} $\uparrow$ &
\textbf{Prec.} $\uparrow$ &
\textbf{Recall} $\uparrow$ &
\textbf{AUC-PR} $\uparrow$ &
\textbf{Prec.} $\uparrow$ &
\textbf{Recall} $\uparrow$ &
\textbf{AUC-PR} $\uparrow$ &
\textbf{Prec.} $\uparrow$ &
\textbf{Recall} $\uparrow$ \\
\midrule

Persistence
& 0.003 & 0.000 & 0.000
& 0.003 & 0.000 & 0.000
& 0.002 & 0.000 & 0.000 \\

Dilation
& 0.164 & 0.051 & 0.274
& 0.127 & 0.059 & 0.191
& 0.144 & 0.058 & 0.227 \\

Cellular Automata
& 0.165 & 0.058 & 0.268
& 0.129 & 0.067 & 0.186
& 0.145 & 0.063 & 0.224 \\

\midrule

U-Net
& \meanstd{0.285}{0.007}
& \meanstd{0.264}{0.008}
& \bestmeanstd{0.378}{0.017}
& \meanstd{0.218}{0.003}
& \meanstd{0.248}{0.042}
& \bestmeanstd{0.272}{0.044}
& \meanstd{0.175}{0.007}
& \meanstd{0.194}{0.044}
& \meanstd{0.247}{0.037} \\

\makecell[l]{U-Net+Time-Gap \\Encoding}
& \bestmeanstd{0.287}{0.007}
& \meanstd{0.268}{0.018}
& \meanstd{0.377}{0.030}
& \bestmeanstd{0.233}{0.014}
& \meanstd{0.283}{0.008}
& \meanstd{0.258}{0.034}
& \bestmeanstd{0.192}{0.011}
& \meanstd{0.220}{0.005}
& \meanstd{0.250}{0.016} \\

Attention U-Net
& \meanstd{0.280}{0.024}
& \meanstd{0.258}{0.011}
& \bestmeanstd{0.378}{0.020}
& \meanstd{0.199}{0.067}
& \meanstd{0.251}{0.027}
& \meanstd{0.262}{0.041}
& \meanstd{0.155}{0.075}
& \meanstd{0.171}{0.028}
& \bestmeanstd{0.266}{0.032} \\

SegFormer U-Net
& \meanstd{0.205}{0.017}
& \meanstd{0.220}{0.011}
& \meanstd{0.301}{0.040}
& \meanstd{0.181}{0.004}
& \meanstd{0.218}{0.022}
& \meanstd{0.241}{0.033}
& \meanstd{0.162}{0.017}
& \meanstd{0.173}{0.025}
& \meanstd{0.252}{0.054} \\

Swin-U-Net
& \meanstd{0.185}{0.018}
& \meanstd{0.197}{0.033}
& \meanstd{0.320}{0.027}
& \meanstd{0.148}{0.014}
& \meanstd{0.179}{0.032}
& \meanstd{0.259}{0.034}
& \meanstd{0.102}{0.006}
& \meanstd{0.130}{0.019}
& \meanstd{0.213}{0.022} \\

U-Mamba
& \meanstd{0.274}{0.012}
& \bestmeanstd{0.308}{0.024}
& \meanstd{0.369}{0.022}
& \meanstd{0.211}{0.012}
& \bestmeanstd{0.334}{0.013}
& \meanstd{0.228}{0.018}
& \meanstd{0.184}{0.021}
& \bestmeanstd{0.261}{0.011}
& \meanstd{0.210}{0.006} \\

OlmoEarth
& \meanstd{0.277}{0.011}
& \meanstd{0.266}{0.013}
& \meanstd{0.364}{0.020}
& \meanstd{0.221}{0.009}
& \meanstd{0.282}{0.004}
& \meanstd{0.241}{0.013}
& \meanstd{0.183}{0.007}
& \meanstd{0.225}{0.012}
& \meanstd{0.236}{0.035} \\

DINOv3
& \meanstd{0.238}{0.008}
& \meanstd{0.283}{0.022}
& \meanstd{0.307}{0.015}
& \meanstd{0.194}{0.010}
& \meanstd{0.259}{0.004}
& \meanstd{0.212}{0.012}
& \meanstd{0.189}{0.006}
& \meanstd{0.224}{0.008}
& \meanstd{0.234}{0.024} \\

\bottomrule
\end{tabular}
\end{adjustbox}

\end{table}

\subsection{Benchmark Performance}
Table~\ref{tab:benchmark_results_mono_temporal} presents the performance of mono-temporal models across the three test sets. Overall, deep learning models substantially outperform the non-learning-based baselines (persistence, dilation, and cellular automata), demonstrating the benefits of learning spatial representations from environmental and satellite observations for next-day wildfire spread prediction.

Among CNN-based architectures, U-Net with time-gap encoding achieves the highest AUC-PR across all three test sets ($0.287$, $0.233$, and $0.192$ for Test-Easy, Test-Medium, and Test-Hard, respectively). This consistent improvement over the standard U-Net suggests that explicitly accounting for the temporal gap between satellite observations and prediction targets helps mitigate temporal misalignment. Attention U-Net achieves the highest recall on Test-Hard ($0.266$) and ties with the standard U-Net on Test-Easy ($0.378$), indicating stronger sensitivity to fire-spread pixels, although at the expense of precision.

Among the more advanced architectures, U-Mamba demonstrates competitive performance, achieving the highest precision on Test-Easy ($0.308$), Test-Medium ($0.334$) and Test-Hard ($0.261$), but without consistently improving AUC-PR over the standard U-Net. In contrast, Transformer-based architectures, particularly Swin-U-Net and SegFormer U-Net, generally underperform the CNN-based models. This suggests that increased architectural complexity does not necessarily translate into better predictive performance for this task.

Finally, pretrained foundation models OlmoEarth and DINOv3 demonstrate competitive performance, with OlmoEarth outperforming several specialized architectures and achieving the second-highest AUC-PR on Test-Medium ($0.221$). However, neither model consistently surpasses the standard U-Net or its time-gap variant. These findings suggest that pretrained representations do not automatically provide an advantage for next-day wildfire spread prediction, highlighting an opportunity for future research to better adapt and fine-tune foundation models using DailyWildfireCAN.

Across all model categories, performance generally decreases from Test-Easy to Test-Hard, highlighting the increasing difficulty of generalizing to more challenging wildfire conditions. 

\begin{table}[htbp]
\centering
\caption{\textit{Multi-temporal} benchmark results for next-day wildfire spread prediction on DailyWildfireCAN. Performance metrics are reported as mean $\pm$ standard deviation across three random seeds. The best mean performance for each metric within each test set is highlighted in bold. Models are ranked in ascending order of AUC-PR (worst to best).}
\label{tab:benchmark_results_multi_temporal}

\footnotesize
\setlength{\tabcolsep}{2pt}
\renewcommand{\arraystretch}{1.1}

\begin{adjustbox}{max width=\linewidth}
\begin{tabular}{@{}lccc|ccc|ccc@{}}
\toprule

\textbf{Model} &
\multicolumn{3}{c}{\textbf{Test-Easy}} &
\multicolumn{3}{c}{\textbf{Test-Medium}} &
\multicolumn{3}{c}{\textbf{Test-Hard}} \\
\cmidrule(lr){2-4}
\cmidrule(lr){5-7}
\cmidrule(lr){8-10}

&
\textbf{AUC-PR} $\uparrow$ &
\textbf{Prec.} $\uparrow$ &
\textbf{Recall} $\uparrow$ &
\textbf{AUC-PR} $\uparrow$ &
\textbf{Prec.} $\uparrow$ &
\textbf{Recall} $\uparrow$ &
\textbf{AUC-PR} $\uparrow$ &
\textbf{Prec.} $\uparrow$ &
\textbf{Recall} $\uparrow$ \\
\midrule

Multi-Model
& \meanstd{0.182}{0.150}
& \meanstd{0.204}{0.176}
& \meanstd{0.239}{0.207}
& \meanstd{0.167}{0.134}
& \meanstd{0.213}{0.184}
& \meanstd{0.168}{0.146}
& \meanstd{0.110}{0.096}
& \meanstd{0.162}{0.140}
& \meanstd{0.158}{0.137} \\

Video Swin-U-Net
& \meanstd{0.272}{0.016}
& \meanstd{0.318}{0.009}
& \meanstd{0.369}{0.020}
& \meanstd{0.235}{0.013}
& \meanstd{0.329}{0.008}
& \meanstd{0.279}{0.020}
& \meanstd{0.178}{0.019}
& \meanstd{0.266}{0.006}
& \meanstd{0.229}{0.032} \\

SimVPv2
& \meanstd{0.284}{0.004}
& \meanstd{0.361}{0.009}
& \meanstd{0.365}{0.014}
& \meanstd{0.250}{0.001}
& \meanstd{0.364}{0.017}
& \meanstd{0.281}{0.038}
& \meanstd{0.203}{0.005}
& \meanstd{0.297}{0.027}
& \meanstd{0.252}{0.044} \\

U-Net ConvLSTM
& \meanstd{0.340}{0.016}
& \bestmeanstd{0.363}{0.016}
& \meanstd{0.388}{0.017}
& \meanstd{0.278}{0.018}
& \meanstd{0.354}{0.009}
& \meanstd{0.286}{0.006}
& \meanstd{0.222}{0.012}
& \meanstd{0.272}{0.008}
& \meanstd{0.248}{0.010} \\

U-TAE
& \bestmeanstd{0.352}{0.004}
& \meanstd{0.358}{0.017}
& \bestmeanstd{0.399}{0.016}
& \bestmeanstd{0.306}{0.002}
& \bestmeanstd{0.362}{0.017}
& \bestmeanstd{0.309}{0.021}
& \bestmeanstd{0.259}{0.005}
& \bestmeanstd{0.296}{0.025}
& \bestmeanstd{0.295}{0.016} \\

\bottomrule
\end{tabular}
\end{adjustbox}

\end{table}

Table~\ref{tab:benchmark_results_multi_temporal} presents the performance of multi-temporal models across the three test sets. Overall, multi-temporal architectures achieve higher AUC-PR than their mono-temporal counterparts, highlighting the benefits of incorporating historical observations for next-day wildfire spread prediction.

\begin{figure}[t]
  \centering

  \begin{subfigure}{0.99\textwidth}
    \centering
    \begin{minipage}{0.04\textwidth}
      \centering
      \rotatebox{90}{\textbf{(a) Mono-Temporal: U-Net}}
    \end{minipage}
    \begin{minipage}{0.94\textwidth}
      \centering
      \includegraphics[width=\textwidth]{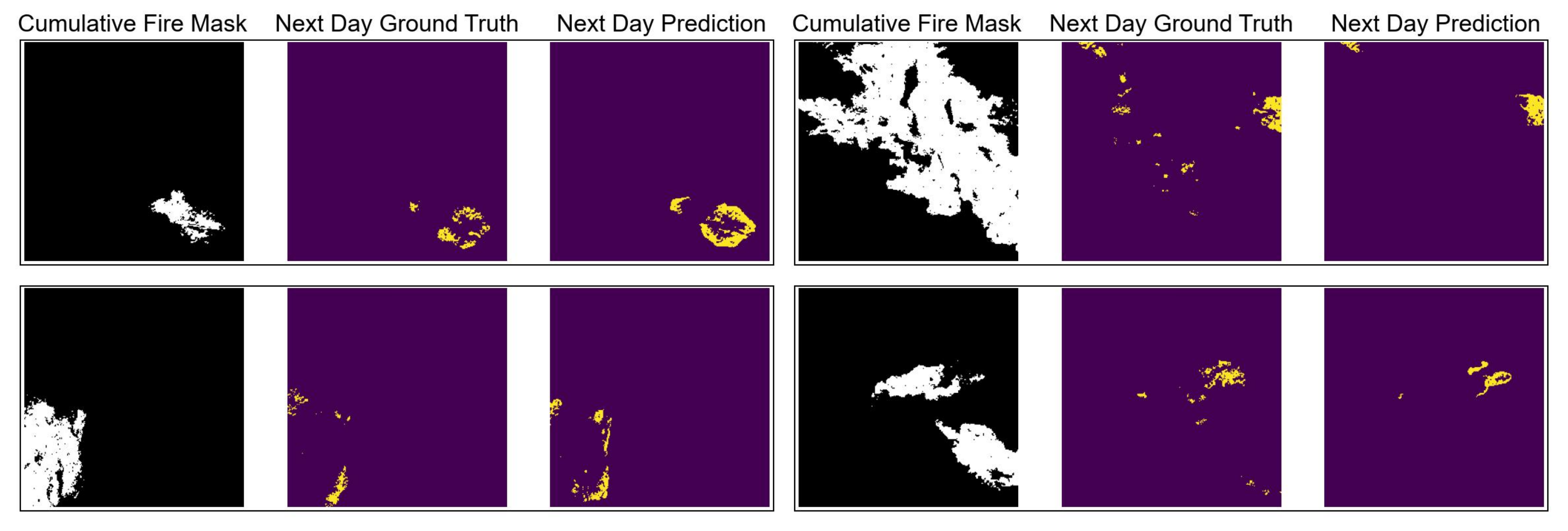}
    \end{minipage}
  \end{subfigure}

  \vspace{0.5em}

  \begin{subfigure}{0.99\textwidth}
    \centering
    \begin{minipage}{0.04\textwidth}
      \centering
      \rotatebox{90}{\textbf{(b) Multi-Temporal: U-TAE}}
    \end{minipage}
    \begin{minipage}{0.94\textwidth}
      \centering
      \includegraphics[width=\textwidth]{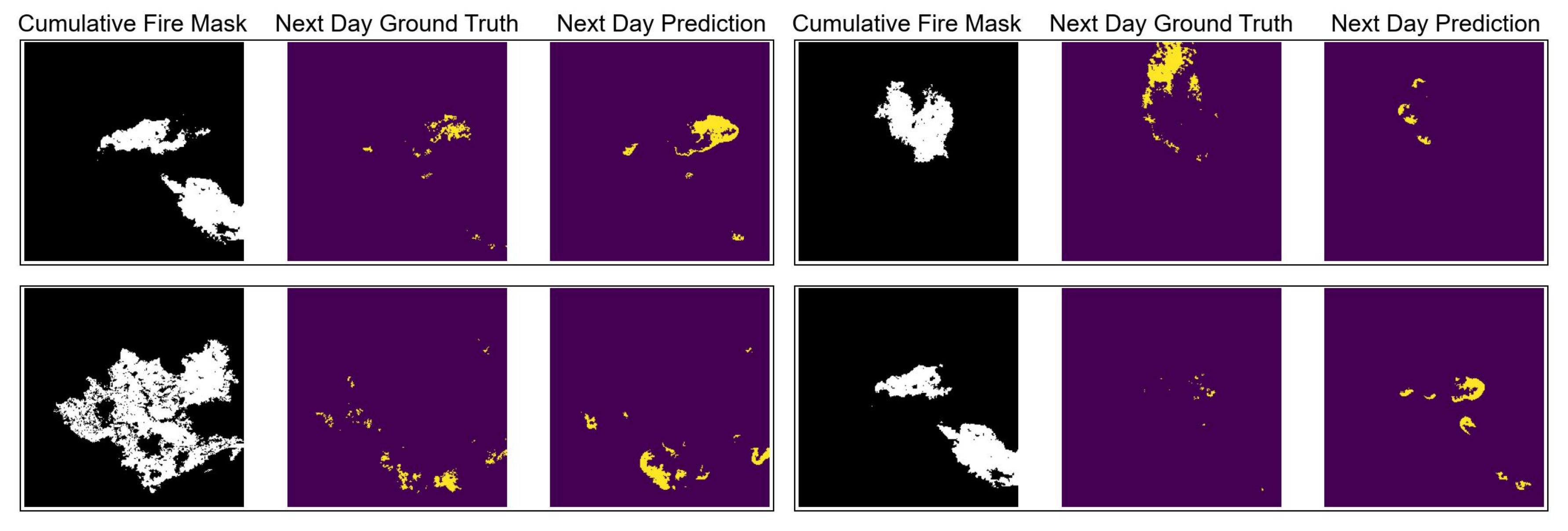}
    \end{minipage}
  \end{subfigure}

  \caption{Qualitative examples of next-day wildfire spread predictions for our two best models: (a) Mono-temporal: U-Net with Time-Gap Encoding and (b) Multi-Temporal: UTAE.}
  \label{fig:sample_predictions}
\end{figure}

Among temporal attention architectures, U-TAE achieves the strongest overall performance, obtaining the highest AUC-PR and recall across all three test sets. Its advantage is particularly evident on Test-Hard, where it achieves an AUC-PR of $0.259$, compared to $0.192$ for the best mono-temporal model, U-Net with time-gap encoding. This suggests that temporal attention can effectively integrate historical observations to improve the detection of wildfire spread, particularly under more challenging conditions. Among recurrent architectures, U-Net with ConvLSTM achieves the second-highest AUC-PR across all test sets and the highest precision on Test-Easy ($0.363$), demonstrating the effectiveness of recurrent temporal modeling. Among the remaining architectures, SimVPv2 consistently outperforms Video Swin-U-Net in AUC-PR but neither surpasses U-TAE or U-Net with ConvLSTM, suggesting that more complex architectures do not necessarily improve wildfire spread prediction. Finally, the multi-branch fusion architecture exhibits the weakest overall performance and high variability across random seeds. This suggests that processing input features through separate branches does not necessarily outperform learning joint representations.

Overall, these results demonstrate the value of multi-temporal modeling, with U-TAE achieving the highest AUC-PR on Test-Easy ($0.352$), Test-Medium ($0.306$), and Test-Hard ($0.259$). Despite these improvements, performance decreases as test difficulty increases, highlighting the need for future research to improve model robustness and generalization under more challenging wildfire spread conditions. Figure~\ref{fig:sample_predictions} provides qualitative examples comparing predictions from mono-temporal and multi-temporal models.  We provide additional input-level ablation studies in ~\ref{app:feature_ablation_results} to assess the predictive contribution of different feature groups and examine how their performance varies when generalizing to unseen years and geographic regions.

\subsection{Robustness and Generalization Analysis}

Beyond aggregate benchmark performance, we investigate how model performance varies across fire characteristics, temporal conditions, and geographic regions. We focus on two representative mono-temporal models, U-Net with Time-Gap encoding (U-Net for short) and OlmoEarth. Fire characteristics are evaluated on \textit{Test-Hard} across four size categories: Extra-Small ($<$600 ha), Small (600--5,000 ha), Medium (5,000--20,000 ha), and Large ($>$20,000 ha), as well as four fire duration stages: Early stage (1-3 days), Developing stage (4-7 days), Established stage (8-14 days), and Late stage (>14 days). We conduct these analyses on different test sets to ensure sufficient and representative samples for each evaluation dimension, while capturing variability across fire characteristics, temporal conditions, and geographic regions. Temporal variability is examined across months on \textit{Test-Hard} and evaluation years on \textit{Test-Medium}. Geographic variability is assessed across Canadian provinces and territories using the union of \textit{Test-Easy} and \textit{Test-Hard}. 

\paragraph{Fire Characteristics - Table~\ref{tab:generalization_analysis_hard}} Both models exhibit decreasing AUC-PR from small to large fires, suggesting greater difficulty in predicting the spread of larger wildfires. Extra-small and small fires achieve comparable AUC-PR, although the extra-small category is constrained by the filtering threshold ($>500$ ha) in our dataset. Across fire duration stages, performance varies moderately during early, developing, and established stages, but declines for late-stage fires, particularly in recall. These findings suggest that predicting wildfire spread becomes more challenging as fires increase in size and duration.

\begin{table}[htbp]
\centering
\caption{Generalization analysis on Test-Hard across fire size, and fire duration. Values are reported as mean
$\pm$ standard deviation over three random seeds.
The higher mean between models is highlighted in bold.}
\label{tab:generalization_analysis_hard}

\footnotesize
\setlength{\tabcolsep}{3pt}
\renewcommand{\arraystretch}{1.1}

\begin{adjustbox}{max width=\linewidth}
\begin{tabular}{@{}lccc@{\hspace{8pt}}ccc@{}}
\toprule
\textbf{Criteria} &
\multicolumn{3}{c}{\textbf{U-Net}} &
\multicolumn{3}{c}{\textbf{OlmoEarth}} \\
\cmidrule(lr){2-4}
\cmidrule(lr){5-7}
&
\textbf{AUC-PR} $\uparrow$ &
\textbf{Prec.} $\uparrow$ &
\textbf{Recall} $\uparrow$ &
\textbf{AUC-PR} $\uparrow$ &
\textbf{Prec.} $\uparrow$ &
\textbf{Recall} $\uparrow$ \\
\midrule

\multicolumn{7}{@{}l}{\textbf{Fire Size}} \\
\addlinespace[2pt]

Extra Small
& \meanstd{0.195}{0.143}
& \meanstd{0.230}{0.156}
& \meanstd{0.203}{0.220}
& \meanstd{0.198}{0.133}
& \meanstd{0.248}{0.125}
& \meanstd{0.197}{0.200} \\

Small
& \meanstd{0.196}{0.009}
& \meanstd{0.212}{0.046}
& \meanstd{0.260}{0.038}
& \meanstd{0.201}{0.020}
& \meanstd{0.241}{0.025}
& \meanstd{0.250}{0.020} \\

Medium
& \meanstd{0.166}{0.016}
& \meanstd{0.146}{0.043}
& \meanstd{0.291}{0.056}
& \meanstd{0.181}{0.031}
& \meanstd{0.185}{0.003}
& \meanstd{0.287}{0.094} \\

Large
& \meanstd{0.160}{0.003}
& \meanstd{0.200}{0.042}
& \meanstd{0.211}{0.045}
& \meanstd{0.159}{0.007}
& \meanstd{0.218}{0.004}
& \meanstd{0.197}{0.033} \\

\midrule
\multicolumn{7}{@{}l}{\textbf{Fire Duration}} \\
\addlinespace[2pt]

Early (1--3 days)
& \meanstd{0.213}{0.009}
& \meanstd{0.215}{0.050}
& \meanstd{0.298}{0.047}
& \meanstd{0.223}{0.023}
& \meanstd{0.254}{0.030}
& \meanstd{0.291}{0.031} \\

Developing (4--7 days)
& \meanstd{0.173}{0.005}
& \meanstd{0.155}{0.024}
& \meanstd{0.299}{0.043}
& \meanstd{0.181}{0.004}
& \meanstd{0.182}{0.014}
& \meanstd{0.290}{0.047} \\

Established (8--14 days)
& \meanstd{0.196}{0.009}
& \meanstd{0.183}{0.048}
& \meanstd{0.298}{0.040}
& \meanstd{0.202}{0.007}
& \meanstd{0.212}{0.013}
& \meanstd{0.292}{0.030} \\

Late Stage ($>$14 days)
& \meanstd{0.163}{0.012}
& \meanstd{0.209}{0.051}
& \meanstd{0.207}{0.035}
& \meanstd{0.172}{0.007}
& \meanstd{0.238}{0.008}
& \meanstd{0.195}{0.037} \\

\bottomrule
\end{tabular}
\end{adjustbox}
\end{table}

\paragraph{Geographic Variability - Table~\ref{tab:generalization_analysis_provinces}} Regional performance also varies considerably, with both models achieving their lowest AUC-PR in Alberta and their highest in Manitoba, followed by relatively strong performance in Yukon. This regional variation is consistent with fire-size distribution across provinces shown in ~\ref{app:dataset_analysis}, Figure~\ref{fig:fire_sizes_provinces}: provinces with larger fires tend to exhibit lower AUC-PR, aligning with our earlier findings that current models struggle to predict the spread of larger fires. U-Net generally achieves higher recall, whereas OlmoEarth tends to achieve higher precision across provinces and territories. The similar geographic performance patterns of both architectures suggest that regional differences are not specific to a single model. Overall, these analyses reveal substantial performance variability across wildfire characteristics, temporal conditions, and geographic regions, motivating further research into improving model robustness and generalization. 

\begin{table}[htbp]
\centering
\caption{Geographic performance across Canadian provinces and territories on the combined Test-Easy and Test-Hard sets. Regions are ordered by ascending average AUC-PR across U-Net and OlmoEarth. Results are reported as mean $\pm$ standard deviation over three random seeds.}
\label{tab:generalization_analysis_provinces}

\footnotesize
\setlength{\tabcolsep}{3pt}
\renewcommand{\arraystretch}{1.1}

\begin{adjustbox}{max width=\linewidth}
\begin{tabular}{@{}lccc@{\hspace{8pt}}ccc@{}}
\toprule
\textbf{Province/Territory} &
\multicolumn{3}{c}{\textbf{U-Net}} &
\multicolumn{3}{c}{\textbf{OlmoEarth}} \\
\cmidrule(lr){2-4}
\cmidrule(lr){5-7}
&
\textbf{AUC-PR} $\uparrow$ &
\textbf{Prec.} $\uparrow$ &
\textbf{Recall} $\uparrow$ &
\textbf{AUC-PR} $\uparrow$ &
\textbf{Prec.} $\uparrow$ &
\textbf{Recall} $\uparrow$ \\
\midrule

Alberta
& \meanstd{0.192}{0.005}
& \meanstd{0.199}{0.020}
& \meanstd{0.271}{0.026}
& \meanstd{0.187}{0.007}
& \meanstd{0.206}{0.005}
& \meanstd{0.258}{0.0200} \\

Northwest Territories
& \meanstd{0.229}{0.006}
& \meanstd{0.250}{0.012}
& \meanstd{0.286}{0.022}
& \meanstd{0.220}{0.010}
& \meanstd{0.250}{0.017}
& \meanstd{0.273}{0.027} \\

British Columbia
& \meanstd{0.273}{0.003}
& \meanstd{0.221}{0.010}
& \meanstd{0.405}{0.013}
& \meanstd{0.267}{0.012}
& \meanstd{0.226}{0.014}
& \meanstd{0.400}{0.016} \\

Quebec
& \meanstd{0.298}{0.015}
& \meanstd{0.258}{0.030}
& \meanstd{0.412}{0.008}
& \meanstd{0.299}{0.013}
& \meanstd{0.276}{0.005}
& \meanstd{0.401}{0.006} \\

Ontario
& \meanstd{0.301}{0.012}
& \meanstd{0.287}{0.029}
& \meanstd{0.382}{0.025}
& \meanstd{0.302}{0.013}
& \meanstd{0.315}{0.012}
& \meanstd{0.356}{0.003} \\

Saskatchewan
& \meanstd{0.309}{0.011}
& \meanstd{0.289}{0.008}
& \meanstd{0.394}{0.022}
& \meanstd{0.301}{0.011}
& \meanstd{0.292}{0.013}
& \meanstd{0.378}{0.022} \\

Yukon
& \meanstd{0.334}{0.008}
& \meanstd{0.274}{0.003}
& \meanstd{0.460}{0.015}
& \meanstd{0.327}{0.012}
& \meanstd{0.273}{0.009}
& \meanstd{0.452}{0.009} \\

Manitoba
& \meanstd{0.357}{0.006}
& \meanstd{0.315}{0.016}
& \meanstd{0.451}{0.029}
& \meanstd{0.347}{0.012}
& \meanstd{0.328}{0.026}
& \meanstd{0.420}{0.013} \\

\bottomrule
\end{tabular}
\end{adjustbox}
\end{table}

\paragraph{Temporal Variability} Analysis shows that both models exhibit substantial seasonal and interannual performance variability, achieving their highest monthly AUC-PR in April and their highest yearly AUC-PR in 2018. These findings highlight the importance of evaluating wildfire spread prediction across diverse temporal conditions, with detailed results provided in ~\ref{app:additional_temporal_analysis}. We further assess geographic and temporal generalization through additional cross-region experiments using DINOv3, with results reported in~\ref{app:cross_region_testing}.

\subsection{Spatial Prediction Quality} 
Beyond conventional pixel-level metrics, we evaluate spatial prediction quality using two complementary measures:

\begin{itemize}
\item \textbf{Boundary Accuracy (F1):} Measures the alignment between predicted and reference fire boundaries under spatial tolerances of $90$, $180$, and $270$ m (1--3 pixels).
\item \textbf{Spread-Distance Performance (F1):} Evaluates prediction performance across five distance bands (0--90, 90--180, 180--270, 270--360, and $>$360 m) extending outward from the previous fire perimeter.
\end{itemize}

We adopt the F1-score for both spatial evaluations, as it balances precision and recall while accounting for the strong class imbalance inherent in wildfire spread prediction. For boundary accuracy, F1 measures the agreement between predicted and reference fire boundaries within a specified spatial tolerance, penalizing both missed and incorrectly predicted boundaries. For spread-distance analysis, it provides a consistent measure of prediction quality across distance bands, capturing the trade-off between detecting fire spread and limiting false positives.

\begin{figure}[htbp]
    \centering

    \begin{subfigure}[t]{0.495\linewidth}
        \centering
        \includegraphics[width=\linewidth]{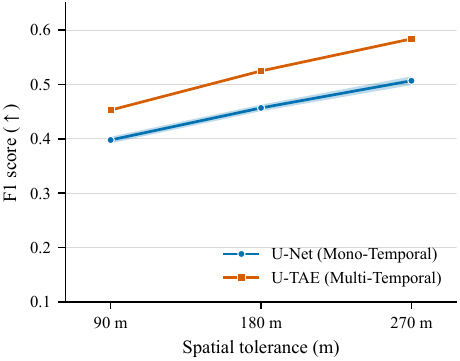}
        \caption{Boundary accuracy across spatial tolerances.}
        \label{fig:boundary_accuracy}
    \end{subfigure}
    \hfill
    \begin{subfigure}[t]{0.495\linewidth}
        \centering
        \includegraphics[width=\linewidth]{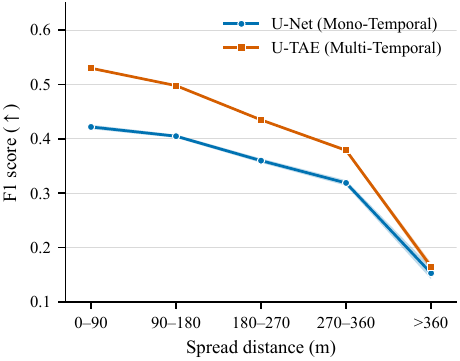}
        \caption{Spread-distance performance across distance bands.}
        \label{fig:spread_distance}
    \end{subfigure}

    \caption{Spatial prediction quality of U-Net (Mono-temporal model) and U-TAE (Multi-temporal model)
    across the combined test sets using all available input
    features. (a) Boundary F1 at spatial tolerances of
    90, 180, and 270 m. (b) Spread-distance F1 across
    five distance bands extending outward from the
    previous fire perimeter. Results are reported as
    mean $\pm$ standard deviation over three random seeds.}

    \label{fig:spatial_analysis}
\end{figure}

Figure~\ref{fig:spatial_analysis} compares U-Net (Mono-temporal model) and U-TAE (Multi-temporal models) (using all input features) across the three test sets combined, while detailed feature-group results are provided in ~\ref{app:additional_spatial_analysis}.

\paragraph{Boundary Accuracy}
Both models exhibit consistent improvements in boundary F1 as spatial tolerance increases from 90 to 270 m. U-Net improves from $0.398$ to $0.507$, while U-TAE improves from $0.453$ to $0.584$, corresponding to absolute gains of $0.109$ and $0.131$, respectively. U-TAE consistently outperforms U-Net across all tolerances, suggesting better spatial localization of predicted wildfire spread. These findings highlight the sensitivity of boundary evaluation to spatial tolerance, which is particularly relevant for applications where approximate fire boundaries still provide useful information for decision-making.

\paragraph{Spread-Distance Performance}
Both architectures exhibit decreasing F1 as spread distance increases. U-Net declines from 0.422 within 0--90 m to 0.153 beyond 360 m, while U-TAE decreases from 0.530 to 0.165. Although U-TAE consistently outperforms U-Net, both models struggle to predict spread farther from the current fire perimeter. This highlights long-distance fire spread prediction as an important direction for future research.

\paragraph{Feature Group Analysis}
The detailed results in Tables~\ref{tab:monotemporal_spatial_analysis_v2} and ~\ref{tab:multitemporal_spatial_analysis_v2} (~\ref{app:additional_spatial_analysis}) show that combining all available features generally provides the strongest spread-distance performance, achieving the highest F1 across the first four distance bands for both models. However, individual feature groups show competitive performance under certain conditions. In particular, land cover achieves the highest F1 beyond 360 m for U-TAE, while land cover and vegetation indices perform competitively in boundary accuracy. These findings suggest that the contribution of different input features varies with the spatial characteristics of fire spread, motivating further investigation into scale-aware feature representations and fusion strategies.

\subsection{Robustness to Spatial and Temporal Label Variations}
Beyond conventional evaluation, DailyWildfireCAN enables the assessment of model robustness to spatial and temporal variations in reference fire progression labels. Satellite-derived wildfire progression data, such as CFSDS \citep{CFSDS}, combine final burned-area information with MODIS and VIIRS active-fire detections to estimate daily fire progression. As with other satellite-derived observations, these estimates are subject to inherent spatial and temporal uncertainty arising from sensor resolution, satellite overpass timing, cloud and smoke obscuration, and interpolation between observations.

To investigate model sensitivity to these variations, we re-evaluate fixed predictions from U-Net and OlmoEarth mono-temporal models under two alternative target configurations: (1) a one-pixel spatial dilation to assess sensitivity to small boundary displacements, and (2) a three-day merged target $(t-1, t, t+1)$ to assess sensitivity to temporal alignment. Table~\ref{tab:label_sensitivity_analysis} presents the resulting AUC-PR scores across the combined test sets.

Spatial dilation increases AUC-PR from $0.246$ to $0.265$ for U-Net and from $0.243$ to $0.266$ for OlmoEarth, suggesting that small spatial differences between predictions and reference fire boundaries influence evaluation outcomes. In contrast, the three-day temporal target yields lower AUC-PR values of 0.219 and 0.220, respectively. Both models exhibit similar sensitivity to these modifications, indicating that their relative performance remains broadly consistent across evaluation settings. This analysis demonstrates the value of DailyWildfireCAN in supporting uncertainty-aware evaluation and encourages future research into models that are robust to spatial and temporal variations in wildfire progression observations.

\begin{table}[htbp]
\centering
\caption{Sensitivity of model performance to spatial and temporal variations in reference fire progression labels across the combined test sets. Results are reported as mean $\pm$ standard deviation over three random seeds.}
\label{tab:label_sensitivity_analysis}

\footnotesize
\setlength{\tabcolsep}{4pt}
\renewcommand{\arraystretch}{1.15}

\begin{adjustbox}{max width=\linewidth}
\begin{tabular}{@{}lccc@{}}
\toprule
\textbf{Model} &
\textbf{Original} &
\textbf{Temporal} &
\textbf{Spatial} \\
&
\textbf{AUC-PR} $\uparrow$ &
\textbf{AUC-PR} $\uparrow$ &
\textbf{AUC-PR} $\uparrow$ \\
\midrule
U-Net
& $0.246 \pm 0.004$
& $0.219 \pm 0.008$
& $0.265 \pm 0.007$ \\

OlmoEarth
& $0.243 \pm 0.007$
& $0.220 \pm 0.008$
& $0.266 \pm 0.005$ \\
\bottomrule
\end{tabular}
\end{adjustbox}
\end{table}

Collectively, these evaluations highlight the importance of assessing models beyond aggregate performance metrics and the challenges of next-day wildfire spread prediction. Despite improvements from multi-temporal architectures, generalization remains a key limitation, with substantial performance variability across fire characteristics, geographic regions, and temporal conditions. Predicting wildfire spread at $90$ m resolution over a $24$-hour horizon is inherently challenging, particularly given the complex interactions between environmental factors and fire dynamics. Nevertheless, spatial analyses show that prediction quality improves when allowing larger boundary tolerances, suggesting that some errors reflect spatial misalignment rather than a complete failure to capture fire spread patterns. These findings motivate the development of more robust and generalizable spatio-temporal models that can better address these challenges. By exposing these limitations through systematic evaluation, DailyWildfireCAN provides a strong foundation for advancing next-day wildfire spread prediction.

%% file: sections/conclusion.tex
We present DailyWildfireCAN, a large-scale, multi-modal, multi-temporal dataset and benchmark for next-day wildfire spread prediction across Canada. 
The dataset provides daily fire-spread labels at \(90\) m grid resolution and integrates satellite imagery with vegetation, land cover, forest structure, topography, weather features, and human influence to support spatial prediction of wildfire propagation.

Through a comprehensive benchmark of mono-temporal and multi-temporal architectures, we evaluate the ability of deep learning models to predict wildfire spread from multimodal geospatial observations and demonstrate the benefits of incorporating historical information.
Beyond aggregate predictive performance, our multi-dimensional evaluation framework reveals substantial variability across fire characteristics, temporal conditions, and geographic regions, while providing additional insights into spatial prediction quality and sensitivity to reference label variations. These analyses highlight the importance of evaluating wildfire spread models across diverse conditions and spatial scales.

The findings also identify promising directions for future research, including improved modeling of large and long-duration fires, enhanced geographic generalization, more accurate boundary localization and long-distance spread prediction, longer historical input sequences, and improved adaptation of pretrained foundation models to wildfire-specific tasks. Future dataset releases could expand temporal coverage, incorporate additional features from existing modalities, and integrate new data sources. 
Important technical limitations remain, including coarse weather inputs, and the restriction to already active fires (as further discussed in~\ref{appendix:limitations}). Together, DailyWildfireCAN dataset, benchmark and the evaluation framework provide a foundation for developing and assessing generalizable and more robust wildfire spread prediction models.

\color{black}

%% file: sections/appendix.tex
\newpage
\section{Further Dataset Analysis}
\label{app:dataset_analysis}

\begin{figure*}
  \centering
  \includegraphics[width=0.99\textwidth]{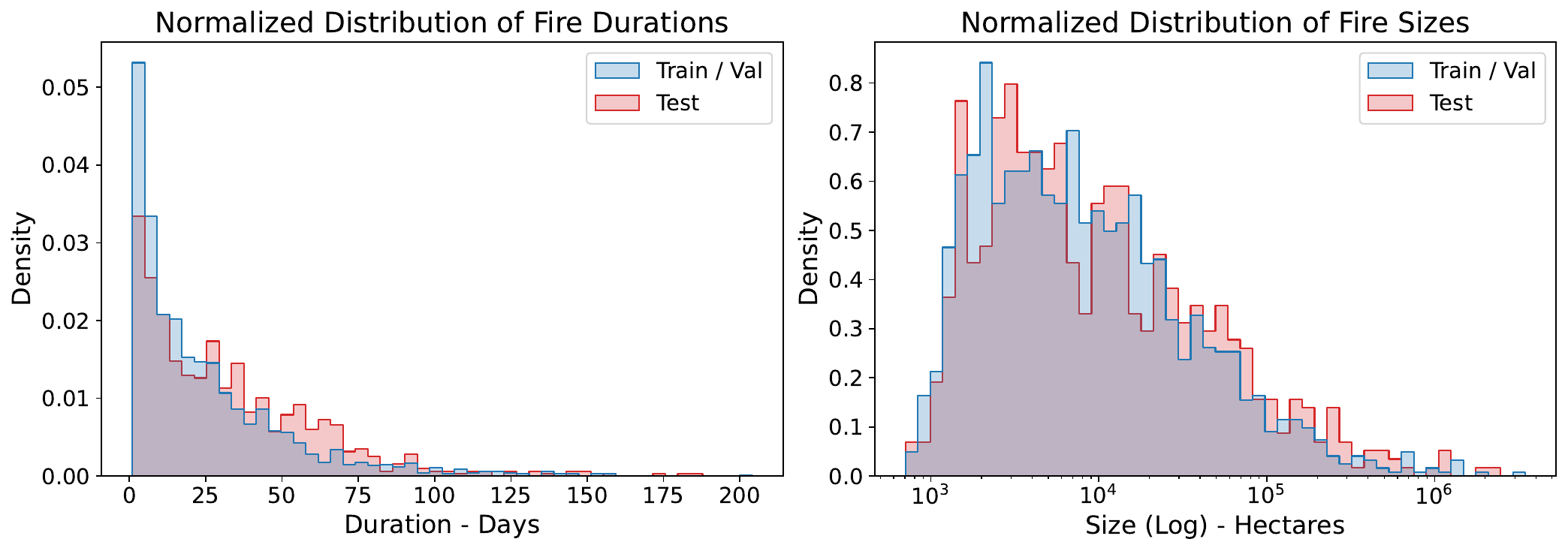}
  \caption{Distribution of the fire sizes and fire durations across the training, validation, and test splits.}
  \label{fig:dataset_histograms}
\end{figure*}

\begin{figure*}
  \centering
  \includegraphics[width=0.99\textwidth]{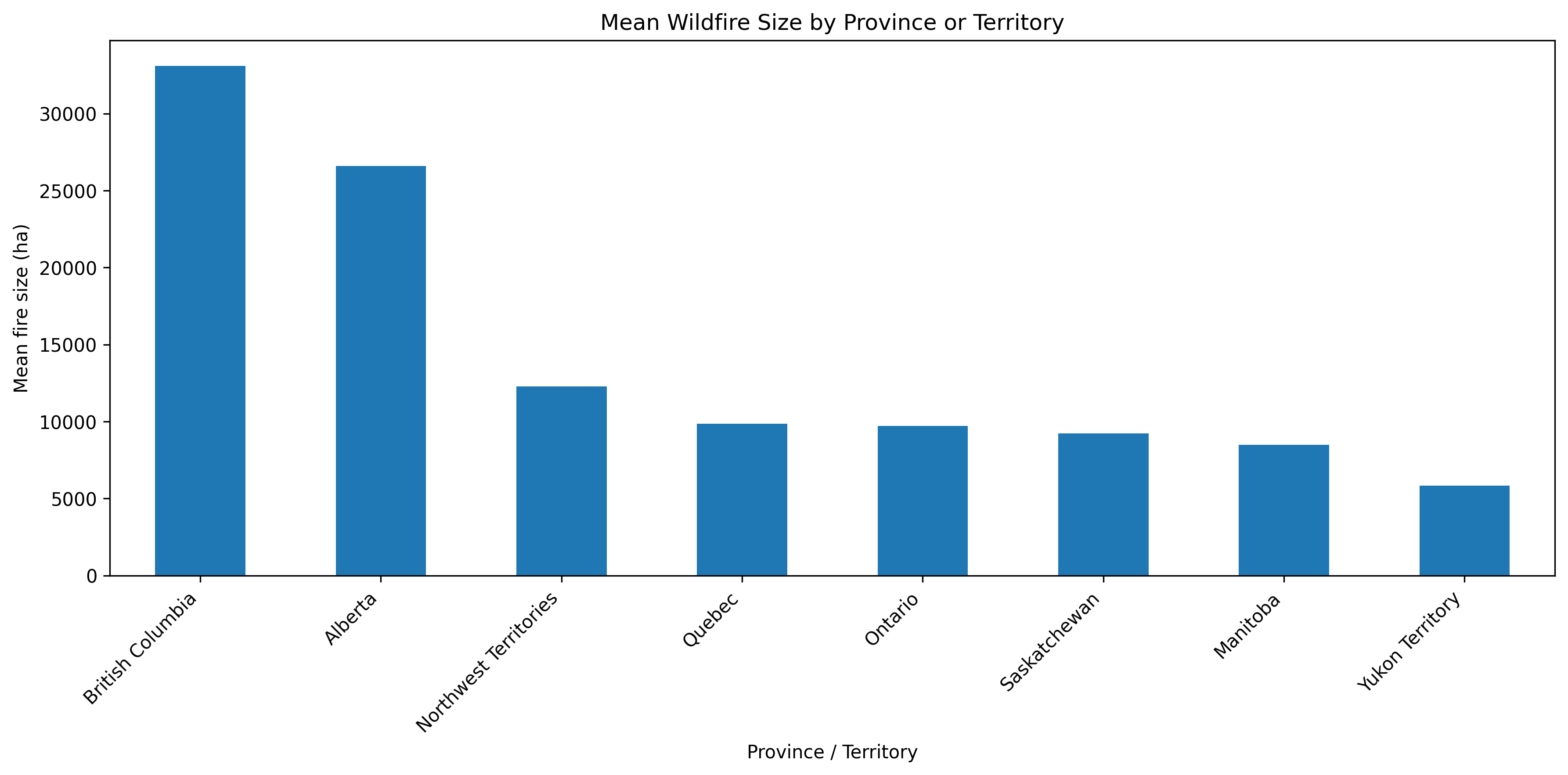}
  \caption{Mean fire sizes in hectares across provinces for the training samples.}
  \label{fig:fire_sizes_provinces}
\end{figure*}

Figure \ref{fig:dataset_histograms} shows the histogram distribution of fire duration in days and fire sizes in hectares across the three splits. Overall, the splits show highly similar distributions, suggesting that the dataset partitioning preserves the main statistical properties of the full dataset.

Figure \ref{fig:fire_sizes_provinces} illustrates the mean fire size (in hectares) across training samples by province. Alberta and British Columbia exhibit larger average fire sizes, while Manitoba and the Yukon Territory show comparatively smaller averages.

\section{Benchmark Training Details} \label{appendix:training_details}

We train all our models for $20$ epochs, using the AdamW optimizer \citep{loshchilov2018_adamW}. We use a weighted loss combining focal loss \citep{focal_loss} and Dice Loss \citep{dice_loss}, equally weighted at 0.5 each. We used a batch size of $16$, with an initial learning rate of $0.001$ and weight decay of $0.0001$ with a cosine learning rate scheduler. All model configurations were trained on a single NVIDIA H100 GPU with a $32$ GB memory allocation and 4 CPU cores per task. Depending on the specific hardware node, individual training runs required between four (for mono-temporal models) and twelve hours (for multi-temporal models). All models are trained over $3$ different runs; each run has its own seed that changes both training initialization and data shuffling. Best checkpoints were selected based on the best loss value on the validation set; the final set of hyperparameters was selected the same way.\\
Finally, we used the pretrained OlmoEarth-v1.2-Base and DINOv3 models as frozen feature extractors, using Sentinel-2 spectral bands as inputs to generate embeddings. The remaining input features were processed by a U-Net encoder, and the resulting feature representations were concatenated with the pretrained embeddings at the bottleneck. During training, the concatenated feature representations were processed jointly by the decoder, which was trained to produce the final predictions.

\section{Additional Experiments and Analysis} \label{appendix:additional_experiments}

\subsection{Feature Ablation} \label{app:feature_ablation_results}
We conduct additional experiments to assess the predictive value of different input feature groups. All ablation experiments reported in Tables \ref{tab:feature_ablation} and \ref{tab:fire_mask_ablation} use a U-Net trained for $20$ epochs across three random seeds. In the first ablation, we train a separate model using each feature group individually, while keeping the accumulated fire mask as an input.

Table \ref{tab:feature_ablation} shows that the relative importance of feature groups varies across the three test sets. On Test-Easy, SCANFI forest structure features achieve the highest AUC-PR and Recall, while satellite imagery yields the highest Precision. On Test-Medium, vegetation indices (NDVI and EVI) achieve the highest AUC-PR and Recall, whereas satellite imagery provides the highest Precision. On Test-Hard, satellite imagery achieves the highest AUC-PR and Precision, while vegetation indices yield the highest Recall. These results suggest that the predictive value of individual feature groups varies across temporal and geographic generalization settings, with forest structure performing best in previously seen regions and satellite imagery and vegetation indices showing stronger performance under more challenging evaluation conditions of unseen regions.

Table \ref{tab:fire_mask_ablation} further investigates the contribution of the accumulated fire mask. Removing this input leads to a substantial performance degradation, highlighting the importance of the current fire state for predicting next-day wildfire spread.

\begin{table}[htbp]
\centering
\caption{Ablation study on the importance of different groups of input features. Metrics are reported on the test sets as mean $\pm$ standard deviation across three random seeds.}
\label{tab:feature_ablation}

\small
\setlength{\tabcolsep}{2pt}
\renewcommand{\arraystretch}{1.1}

\begin{adjustbox}{width=\linewidth}
\begin{tabular}{@{}lccc|ccc|ccc@{}}
\toprule

\textbf{Feature Group} &
\multicolumn{3}{c}{\textbf{Test-Easy}} &
\multicolumn{3}{c}{\textbf{Test-Medium}} &
\multicolumn{3}{c}{\textbf{Test-Hard}} \\
\cmidrule(lr){2-4}
\cmidrule(lr){5-7}
\cmidrule(lr){8-10}

&
\textbf{AUC-PR} $\uparrow$ &
\textbf{Prec.} $\uparrow$ &
\textbf{Recall} $\uparrow$ &
\textbf{AUC-PR} $\uparrow$ &
\textbf{Prec.} $\uparrow$ &
\textbf{Recall} $\uparrow$ &
\textbf{AUC-PR} $\uparrow$ &
\textbf{Prec.} $\uparrow$ &
\textbf{Recall} $\uparrow$ \\
\midrule

Dilation Baseline
& 0.164 & 0.051 & 0.274
& 0.127 & 0.059 & 0.191
& 0.144 & 0.058 & 0.227 \\
\midrule
\midrule

Vegetation (NDVI \& EVI)
& \meanstd{0.206}{0.019}
& \meanstd{0.156}{0.011}
& \meanstd{0.390}{0.016}
& \meanstd{0.171}{0.024}
& \meanstd{0.185}{0.010}
& \meanstd{0.285}{0.026}
& \meanstd{0.183}{0.015}
& \meanstd{0.145}{0.021}
& \meanstd{0.348}{0.022} \\
\midrule

Forest Features (SCANFI)
& \meanstd{0.243}{0.008}
& \meanstd{0.145}{0.011}
& \meanstd{0.431}{0.019}
& \meanstd{0.153}{0.032}
& \meanstd{0.134}{0.063}
& \meanstd{0.253}{0.025}
& \meanstd{0.123}{0.021}
& \meanstd{0.097}{0.054}
& \meanstd{0.227}{0.006} \\
\midrule

Land Cover
& \meanstd{0.210}{0.009}
& \meanstd{0.168}{0.013}
& \meanstd{0.348}{0.022}
& \meanstd{0.157}{0.014}
& \meanstd{0.156}{0.009}
& \meanstd{0.247}{0.033}
& \meanstd{0.186}{0.003}
& \meanstd{0.132}{0.009}
& \meanstd{0.339}{0.005} \\
\midrule

Disturbance (CanLaD)
& \meanstd{0.171}{0.021}
& \meanstd{0.127}{0.004}
& \meanstd{0.351}{0.029}
& \meanstd{0.132}{0.009}
& \meanstd{0.146}{0.006}
& \meanstd{0.239}{0.014}
& \meanstd{0.164}{0.012}
& \meanstd{0.124}{0.009}
& \meanstd{0.331}{0.027} \\
\midrule

Topography
& \meanstd{0.173}{0.012}
& \meanstd{0.132}{0.008}
& \meanstd{0.340}{0.036}
& \meanstd{0.144}{0.007}
& \meanstd{0.188}{0.027}
& \meanstd{0.207}{0.017}
& \meanstd{0.131}{0.020}
& \meanstd{0.150}{0.004}
& \meanstd{0.224}{0.015} \\
\midrule

Weather
& \meanstd{0.174}{0.006}
& \meanstd{0.163}{0.007}
& \meanstd{0.323}{0.026}
& \meanstd{0.147}{0.006}
& \meanstd{0.168}{0.009}
& \meanstd{0.250}{0.012}
& \meanstd{0.174}{0.001}
& \meanstd{0.143}{0.017}
& \meanstd{0.327}{0.017} \\
\midrule

Satellite
& \meanstd{0.199}{0.012}
& \meanstd{0.205}{0.024}
& \meanstd{0.362}{0.038}
& \meanstd{0.157}{0.005}
& \meanstd{0.198}{0.017}
& \meanstd{0.274}{0.033}
& \meanstd{0.188}{0.002}
& \meanstd{0.168}{0.025}
& \meanstd{0.297}{0.041} \\
\midrule

Human Influence Index
& \meanstd{0.157}{0.006}
& \meanstd{0.137}{0.006}
& \meanstd{0.344}{0.008}
& \meanstd{0.123}{0.002}
& \meanstd{0.150}{0.004}
& \meanstd{0.232}{0.012}
& \meanstd{0.154}{0.001}
& \meanstd{0.132}{0.006}
& \meanstd{0.321}{0.010} \\

\bottomrule
\end{tabular}
\end{adjustbox}

\end{table}

\begin{table}[htbp]
\centering
\caption{Ablation study on the importance of the cumulative fire mask as an input feature. Metrics are reported as mean $\pm$ standard deviation across three random seeds. The best mean performance for each metric within each test set is highlighted in bold.}
\label{tab:fire_mask_ablation}

\small
\setlength{\tabcolsep}{2pt}
\renewcommand{\arraystretch}{1.1}

\begin{adjustbox}{width=\linewidth}
\begin{tabular}{@{}lccc|ccc|ccc@{}}
\toprule

\textbf{Model} &
\multicolumn{3}{c}{\textbf{Test-Easy}} &
\multicolumn{3}{c}{\textbf{Test-Medium}} &
\multicolumn{3}{c}{\textbf{Test-Hard}} \\
\cmidrule(lr){2-4}
\cmidrule(lr){5-7}
\cmidrule(lr){8-10}

&
\textbf{AUC-PR} $\uparrow$ &
\textbf{Prec.} $\uparrow$ &
\textbf{Recall} $\uparrow$ &
\textbf{AUC-PR} $\uparrow$ &
\textbf{Prec.} $\uparrow$ &
\textbf{Recall} $\uparrow$ &
\textbf{AUC-PR} $\uparrow$ &
\textbf{Prec.} $\uparrow$ &
\textbf{Recall} $\uparrow$ \\
\midrule

\makecell[l]{All features excluding \\ cumulative fire mask}
& \meanstd{0.014}{0.001}
& \meanstd{0.020}{0.002}
& \meanstd{0.101}{0.010}
& \meanstd{0.019}{0.001}
& \meanstd{0.027}{0.004}
& \meanstd{0.190}{0.032}
& \meanstd{0.010}{0.001}
& \meanstd{0.015}{0.001}
& \meanstd{0.073}{0.053} \\
\midrule

Cumulative fire mask only
& \meanstd{0.162}{0.016}
& \meanstd{0.128}{0.006}
& \meanstd{0.345}{0.023}
& \meanstd{0.122}{0.009}
& \meanstd{0.153}{0.004}
& \meanstd{0.219}{0.023}
& \meanstd{0.160}{0.011}
& \meanstd{0.127}{0.008}
& \bestmeanstd{0.323}{0.013} \\
\midrule

All features
& \bestmeanstd{0.285}{0.007}
& \bestmeanstd{0.264}{0.008}
& \bestmeanstd{0.378}{0.017}
& \bestmeanstd{0.218}{0.003}
& \bestmeanstd{0.248}{0.042}
& \bestmeanstd{0.272}{0.044}
& \bestmeanstd{0.175}{0.007}
& \bestmeanstd{0.194}{0.044}
& \meanstd{0.247}{0.037} \\

\bottomrule
\end{tabular}
\end{adjustbox}

\end{table}

\subsection{Generalization Analysis: Temporal variability} \label{app:additional_temporal_analysis}
We further investigate temporal variability in wildfire spread prediction using U-Net and OlmoEarth, evaluating monthly performance on \textit{Test-Hard} and interannual performance on \textit{Test-Medium}. Table~\ref{tab:temporal_performance_analysis} reports AUC-PR, precision, and recall across both temporal dimensions.

\paragraph{Seasonal Variability.}
Both models achieve their highest monthly AUC-PR in April (0.333 for U-Net and 0.335 for OlmoEarth), while June and September exhibit substantially lower performance. These differences suggest that prediction difficulty varies across the fire season, potentially reflecting changes in environmental conditions, fire behavior, and sample composition.

\paragraph{Interannual Variability.}
Across evaluation years, both models achieve their highest AUC-PR in 2018 (0.272 for U-Net and 0.267 for OlmoEarth), with considerable variability observed across the remaining years. Differences in wildfire activity and fire characteristics, including the relatively high fire activity in 2023 and the prevalence of smaller fires in 2019 and 2020, may contribute to these variations. Overall, these results demonstrate that model performance varies across seasonal and interannual conditions, reinforcing the importance of temporally diverse evaluation in wildfire spread prediction.

\begin{table}[htbp]
\centering
\caption{Temporal performance analysis across months (Test-Hard) and evaluation years (Test-Medium). Results are reported as mean $\pm$ standard deviation over three random seeds.}
\label{tab:temporal_performance_analysis}

\footnotesize
\setlength{\tabcolsep}{3pt}
\renewcommand{\arraystretch}{1.1}

\begin{adjustbox}{max width=\linewidth}
\begin{tabular}{@{}lccc@{\hspace{8pt}}ccc@{}}
\toprule
\textbf{Period} &
\multicolumn{3}{c}{\textbf{U-Net}} &
\multicolumn{3}{c}{\textbf{OlmoEarth}} \\
\cmidrule(lr){2-4}
\cmidrule(lr){5-7}
&
\textbf{AUC-PR} $\uparrow$ &
\textbf{Prec.} $\uparrow$ &
\textbf{Recall} $\uparrow$ &
\textbf{AUC-PR} $\uparrow$ &
\textbf{Prec.} $\uparrow$ &
\textbf{Recall} $\uparrow$ \\

\midrule
\multicolumn{7}{@{}l}{\textbf{Monthly Performance (Test-Hard)}} \\
\midrule

March
& \meanstd{0.159}{0.097}
& \meanstd{0.216}{0.226}
& \meanstd{0.278}{0.050}
& \meanstd{0.211}{0.067}
& \meanstd{0.201}{0.059}
& \meanstd{0.331}{0.170} \\

April
& \meanstd{0.333}{0.044}
& \meanstd{0.197}{0.027}
& \meanstd{0.587}{0.080}
& \meanstd{0.335}{0.032}
& \meanstd{0.234}{0.050}
& \meanstd{0.547}{0.084} \\

June
& \meanstd{0.107}{0.008}
& \meanstd{0.102}{0.033}
& \meanstd{0.227}{0.096}
& \meanstd{0.137}{0.033}
& \meanstd{0.136}{0.010}
& \meanstd{0.254}{0.095} \\

July
& \meanstd{0.157}{0.008}
& \meanstd{0.165}{0.0387}
& \meanstd{0.249}{0.034}
& \meanstd{0.164}{0.007}
& \meanstd{0.194}{0.019}
& \meanstd{0.238}{0.033} \\

August
& \meanstd{0.195}{0.011}
& \meanstd{0.228}{0.052}
& \meanstd{0.247}{0.041}
& \meanstd{0.204}{0.010}
& \meanstd{0.260}{0.007}
& \meanstd{0.234}{0.040} \\

September
& \meanstd{0.115}{0.005}
& \meanstd{0.166}{0.033}
& \meanstd{0.153}{0.036}
& \meanstd{0.123}{0.008}
& \meanstd{0.190}{0.009}
& \meanstd{0.167}{0.014} \\

\midrule
\multicolumn{7}{@{}l}{\textbf{Yearly Performance (Test-Medium)}} \\
\midrule

2017
& \meanstd{0.220}{0.004}
& \meanstd{0.281}{0.029}
& \meanstd{0.222}{0.031}
& \meanstd{0.225}{0.010}
& \meanstd{0.307}{0.010}
& \meanstd{0.202}{0.021} \\

2018
& \meanstd{0.272}{0.003}
& \meanstd{0.298}{0.019}
& \meanstd{0.312}{0.029}
& \meanstd{0.267}{0.007}
& \meanstd{0.311}{0.016}
& \meanstd{0.288}{0.022} \\

2019
& \meanstd{0.154}{0.087}
& \meanstd{0.231}{0.052}
& \meanstd{0.212}{0.182}
& \meanstd{0.181}{0.094}
& \meanstd{0.310}{0.130}
& \meanstd{0.045}{0.039} \\

2020
& \meanstd{0.171}{0.033}
& \meanstd{0.175}{0.092}
& \meanstd{0.280}{0.146}
& \meanstd{0.189}{0.048}
& \meanstd{0.319}{0.067}
& \meanstd{0.144}{0.052} \\

2021
& \meanstd{0.255}{0.007}
& \meanstd{0.226}{0.046}
& \meanstd{0.376}{0.066}
& \meanstd{0.249}{0.006}
& \meanstd{0.252}{0.013}
& \meanstd{0.341}{0.019} \\

2022
& \meanstd{0.193}{0.014}
& \meanstd{0.179}{0.067}
& \meanstd{0.328}{0.113}
& \meanstd{0.185}{0.025}
& \meanstd{0.214}{0.018}
& \meanstd{0.273}{0.039} \\

2023
& \meanstd{0.169}{0.006}
& \meanstd{0.199}{0.051}
& \meanstd{0.239}{0.071}
& \meanstd{0.171}{0.005}
& \meanstd{0.244}{0.020}
& \meanstd{0.193}{0.018} \\

\bottomrule
\end{tabular}
\end{adjustbox}
\end{table}

\subsection{Cross-region Testing} \label{app:cross_region_testing}

We conduct additional experiments using DINOv3 to assess
model generalization across geographic regions and fire
seasons. We consider two complementary evaluation settings:

\begin{itemize}
    \item \textbf{Spatial Cross-Region Testing (2017--2023):}
    We adopt a 10-fold spatial evaluation protocol, where
    eight clusters are used for training, one for validation,
    and one for testing in each fold. Performance is reported
    as the mean across the ten held-out test folds, assessing
    generalization to geographically unseen regions.

    \item \textbf{Temporal Holdout (2024):}
    We evaluate an ensemble of the ten trained models on
    the unseen 2024 fire season to assess generalization
    beyond the training period.
\end{itemize}

Table~\ref{tab:cross_region_evaluation} summarizes the results.
DINOv3 achieves an average AUC-PR of 0.304 across the spatial
test folds and 0.285 on the 2024 temporal holdout.
Recall remains comparable across the two settings
(0.415 and 0.405, respectively).
These findings suggest that the model retains predictive
capability across geographically held-out regions and
an unseen fire season. However, differences in the
evaluation distributions and aggregation strategies
preclude attributing the performance difference solely
to temporal generalization.

\begin{table}[htbp]
\centering
\caption{Cross-region and temporal holdout evaluation. Results are reported as mean $\pm$ standard deviation.}
\label{tab:cross_region_evaluation}

\footnotesize
\setlength{\tabcolsep}{4pt}
\renewcommand{\arraystretch}{1.15}

\begin{adjustbox}{max width=\linewidth}
\begin{tabular}{@{}lccc@{}}
\toprule
\textbf{Evaluation Scenario} &
\textbf{AUC-PR} $\uparrow$ &
\textbf{Precision} $\uparrow$ &
\textbf{Recall} $\uparrow$ \\
\midrule

\makecell[l]{Spatial Cross-Region\\  Testing \\ (2017--2023)}
& $0.304 \pm 0.072$
& $0.307 \pm 0.058$
& $0.415 \pm 0.070$ \\

\makecell[l]{Temporal Holdout \\ (2024)}
& $0.285 \pm 0.046$
& $0.269 \pm 0.018$
& $0.405 \pm 0.023$ \\

\bottomrule
\end{tabular}
\end{adjustbox}
\end{table}

\subsection{Spatial Analysis} \label{app:additional_spatial_analysis}

We provide a detailed analysis of spatial prediction performance across different input feature groups for U-Net and U-TAE. Each configuration combines the previous fire-state information with one additional feature group, while the full-feature configuration includes all available inputs. We evaluate boundary F1 at spatial tolerances of 90, 180, and 270 m, alongside spread-distance F1 across five distance bands extending outward from the previous fire perimeter. Results are reported across the combined test sets as mean $\pm$ standard deviation over three random seeds.

\paragraph{Boundary Accuracy}
Tables~\ref{tab:monotemporal_spatial_analysis_v2} and~\ref{tab:multitemporal_spatial_analysis_v2} show that boundary F1 consistently increases with spatial tolerance across all feature configurations. For U-Net, land cover achieves the highest F1 at 90 m (0.426) and 180 m (0.472), while vegetation indices achieve the highest F1 at 270 m (0.512), slightly outperforming the full-feature configuration. In contrast, U-TAE achieves its highest boundary F1 using all available features across all three tolerances (0.453, 0.525, and 0.584, respectively). These results suggest that the benefit of combining multiple feature groups depends on both the architecture and spatial tolerance.

\paragraph{Spread-Distance Performance}
The full-feature configuration achieves the highest F1 across the first four spread-distance bands for both architectures, highlighting the benefits of combining complementary environmental and satellite observations. However, performance decreases substantially beyond 360 m, where U-TAE with land cover achieves the highest F1 (0.198), compared to 0.165 using all features. For U-Net, the full-feature configuration remains the strongest across all distance bands. These findings suggest that different input representations may be more effective at different spread distances, particularly for longer-range wildfire propagation.

\begin{table}[htbp]
\centering
\caption{Spatial prediction performance of U-Net across input feature groups on the combined test sets. Boundary F1 is evaluated at three spatial tolerances, while spread-distance F1 is measured across five distance bands. All configurations include fire-state information. Results are reported as mean $\pm$ standard deviation over three random seeds.}
\label{tab:monotemporal_spatial_analysis_v2}

\footnotesize
\setlength{\tabcolsep}{3pt}
\renewcommand{\arraystretch}{1.15}

\begin{adjustbox}{max width=\linewidth}
\begin{tabular}{@{}lccc@{\hspace{6pt}}ccccc@{}}
\toprule
\multirow{2}{*}{\textbf{Feature Group}} &
\multicolumn{3}{c}{\textbf{Boundary F1} $\uparrow$} &
\multicolumn{5}{c}{\textbf{Spread-Distance F1} $\uparrow$} \\
\cmidrule(lr){2-4}
\cmidrule(lr){5-9}
&
\textbf{90 m} &
\textbf{180 m} &
\textbf{270 m} &
\textbf{0--90 m} &
\textbf{90--180 m} &
\textbf{180--270 m} &
\textbf{270--360 m} &
\textbf{$>$360 m} \\
\midrule

All Features
& \meanstd{0.398}{0.006}
& \meanstd{0.457}{0.006}
& \meanstd{0.507}{0.008}
& \meanstd{0.422}{0.004}
& \meanstd{0.405}{0.003}
& \meanstd{0.360}{0.004}
& \meanstd{0.319}{0.005}
& \meanstd{0.153}{0.011} \\

Land Cover
& \meanstd{\textbf{0.426}}{0.063}
& \meanstd{\textbf{0.472}}{0.058}
& \meanstd{0.510}{0.054}
& \meanstd{0.199}{0.043}
& \meanstd{0.204}{0.033}
& \meanstd{0.180}{0.023}
& \meanstd{0.149}{0.020}
& \meanstd{0.054}{0.026} \\

Forest Features (SCANFI)
& \meanstd{0.403}{0.024}
& \meanstd{0.450}{0.018}
& \meanstd{0.490}{0.016}
& \meanstd{0.239}{0.003}
& \meanstd{0.238}{0.011}
& \meanstd{0.207}{0.013}
& \meanstd{0.176}{0.009}
& \meanstd{0.069}{0.019} \\

Topography
& \meanstd{0.422}{0.025}
& \meanstd{0.466}{0.016}
& \meanstd{0.503}{0.012}
& \meanstd{0.201}{0.022}
& \meanstd{0.201}{0.023}
& \meanstd{0.172}{0.020}
& \meanstd{0.138}{0.016}
& \meanstd{0.043}{0.021} \\

Disturbance
& \meanstd{0.412}{0.036}
& \meanstd{0.458}{0.039}
& \meanstd{0.497}{0.040}
& \meanstd{0.185}{0.028}
& \meanstd{0.187}{0.037}
& \meanstd{0.163}{0.033}
& \meanstd{0.135}{0.024}
& \meanstd{0.054}{0.015} \\

Human Influence Index
& \meanstd{0.404}{0.031}
& \meanstd{0.453}{0.033}
& \meanstd{0.493}{0.034}
& \meanstd{0.182}{0.035}
& \meanstd{0.177}{0.039}
& \meanstd{0.152}{0.032}
& \meanstd{0.127}{0.024}
& \meanstd{0.062}{0.008} \\

Sentinel-2 Imagery
& \meanstd{0.391}{0.037}
& \meanstd{0.441}{0.039}
& \meanstd{0.482}{0.039}
& \meanstd{0.223}{0.036}
& \meanstd{0.217}{0.036}
& \meanstd{0.187}{0.027}
& \meanstd{0.156}{0.016}
& \meanstd{0.065}{0.022} \\

Veg. Indices
& \meanstd{0.413}{0.027}
& \meanstd{0.468}{0.022}
& \meanstd{\textbf{0.512}}{0.016}
& \meanstd{0.204}{0.052}
& \meanstd{0.189}{0.047}
& \meanstd{0.165}{0.038}
& \meanstd{0.147}{0.030}
& \meanstd{0.080}{0.002} \\

Weather
& \meanstd{0.381}{0.022}
& \meanstd{0.426}{0.018}
& \meanstd{0.465}{0.014}
& \meanstd{0.211}{0.044}
& \meanstd{0.212}{0.044}
& \meanstd{0.186}{0.037}
& \meanstd{0.157}{0.029}
& \meanstd{0.069}{0.014} \\

\bottomrule
\end{tabular}
\end{adjustbox}
\end{table}

\begin{table}[htbp]
\centering
\caption{Spatial prediction performance of U-TAE across input feature groups on the combined test sets. Boundary F1 is evaluated at three spatial tolerances, while spread-distance F1 is measured across five distance bands. All configurations include fire-state information. Results are reported as mean $\pm$ standard deviation over three random seeds.}
\label{tab:multitemporal_spatial_analysis_v2}

\footnotesize
\setlength{\tabcolsep}{3pt}
\renewcommand{\arraystretch}{1.15}

\begin{adjustbox}{max width=\linewidth}
\begin{tabular}{@{}lccc@{\hspace{6pt}}ccccc@{}}
\toprule
\multirow{2}{*}{\textbf{Feature Group}} &
\multicolumn{3}{c}{\textbf{Boundary F1} $\uparrow$} &
\multicolumn{5}{c}{\textbf{Spread-Distance F1} $\uparrow$} \\
\cmidrule(lr){2-4}
\cmidrule(lr){5-9}
&
\textbf{90 m} &
\textbf{180 m} &
\textbf{270 m} &
\textbf{0--90 m} &
\textbf{90--180 m} &
\textbf{180--270 m} &
\textbf{270--360 m} &
\textbf{$>$360 m} \\
\midrule

All Features
& \meanstd{\textbf{0.453}}{0.001}
& \meanstd{\textbf{0.525}}{0.002}
& \meanstd{\textbf{0.584}}{0.002}
& \meanstd{\textbf{0.530}}{0.002}
& \meanstd{\textbf{0.498}}{0.001}
& \meanstd{\textbf{0.435}}{0.001}
& \meanstd{\textbf{0.379}}{0.003}
& \meanstd{0.165}{0.007} \\

Land Cover
& \meanstd{0.425}{0.014}
& \meanstd{0.495}{0.012}
& \meanstd{0.555}{0.011}
& \meanstd{0.445}{0.030}
& \meanstd{0.416}{0.014}
& \meanstd{0.374}{0.001}
& \meanstd{0.337}{0.004}
& \meanstd{\textbf{0.198}}{0.008} \\

Forest Features (SCANFI)
& \meanstd{0.440}{0.003}
& \meanstd{0.513}{0.005}
& \meanstd{0.574}{0.007}
& \meanstd{0.464}{0.020}
& \meanstd{0.447}{0.011}
& \meanstd{0.400}{0.002}
& \meanstd{0.356}{0.007}
& \meanstd{0.176}{0.009} \\

Topography
& \meanstd{0.391}{0.023}
& \meanstd{0.458}{0.026}
& \meanstd{0.516}{0.030}
& \meanstd{0.384}{0.033}
& \meanstd{0.371}{0.012}
& \meanstd{0.337}{0.008}
& \meanstd{0.307}{0.012}
& \meanstd{0.133}{0.029} \\

Disturbance
& \meanstd{0.430}{0.039}
& \meanstd{0.491}{0.033}
& \meanstd{0.546}{0.027}
& \meanstd{0.323}{0.075}
& \meanstd{0.306}{0.056}
& \meanstd{0.290}{0.032}
& \meanstd{0.287}{0.012}
& \meanstd{0.164}{0.041} \\

Human Influence Index
& \meanstd{0.416}{0.034}
& \meanstd{0.478}{0.030}
& \meanstd{0.534}{0.026}
& \meanstd{0.331}{0.060}
& \meanstd{0.318}{0.036}
& \meanstd{0.299}{0.015}
& \meanstd{0.285}{0.005}
& \meanstd{0.162}{0.032} \\

Sentinel-2 Imagery
& \meanstd{0.400}{0.014}
& \meanstd{0.471}{0.016}
& \meanstd{0.535}{0.016}
& \meanstd{0.410}{0.032}
& \meanstd{0.388}{0.016}
& \meanstd{0.351}{0.013}
& \meanstd{0.323}{0.017}
& \meanstd{0.181}{0.011} \\

Veg. Indices
& \meanstd{0.406}{0.012}
& \meanstd{0.474}{0.007}
& \meanstd{0.533}{0.004}
& \meanstd{0.353}{0.041}
& \meanstd{0.355}{0.014}
& \meanstd{0.341}{0.016}
& \meanstd{0.325}{0.016}
& \meanstd{0.159}{0.014} \\

Weather
& \meanstd{0.379}{0.024}
& \meanstd{0.438}{0.030}
& \meanstd{0.490}{0.034}
& \meanstd{0.370}{0.024}
& \meanstd{0.338}{0.046}
& \meanstd{0.264}{0.069}
& \meanstd{0.198}{0.075}
& \meanstd{0.052}{0.029} \\

\bottomrule
\end{tabular}
\end{adjustbox}
\end{table}

\section{Limitations} \label{appendix:limitations}

DailyWildfireCAN has several limitations that should be considered when interpreting benchmark results and using the dataset for future research:

\begin{itemize}
    \item \textbf{Temporal alignment.} Some input sources are not available at daily frequency, requiring observations from different acquisition times to be aligned with each prediction target. This may introduce uncertainty about the exact landscape conditions preceding each target day. This occurs for vegetation indices, and satellite imagery.

    \item \textbf{Weather data spatial resolution.} Weather variables are derived from gridded reanalysis products at approximately $9$ km resolution and resampled to the common $90$ m grid. As a result, they may not fully capture local wind, humidity, and precipitation patterns that strongly influence fire behavior at fine spatial scales.
    
    \item \textbf{Ignition assumption.} The dataset assumes fires are already active and does not include negative samples for ignition prediction.

    \item \textbf{Missing operational factors.} Suppression activities, fuel treatments, firebreaks, and real-time incident decisions are not explicitly modeled.

\end{itemize}

As a result, models trained on DailyWildfireCAN should be interpreted as learning from observed historical spread patterns rather than fully modeling all causal drivers of wildfire propagation.

\section{Released Dataset Sheet} \label{app:data_sheet}
We release both the raw dataset and the processed benchmark samples in mono-temporal and multi-temporal configurations. The raw dataset comprises three main components: fire growth points (CSV), yearly fire metadata (JSON), and fire-specific geospatial data (HDF5). The processed samples are organized into training, validation, and three test subsets, accompanied by training-set normalization statistics. We additionally provide a representative sample and an accompanying notebook to facilitate data exploration. The following subsections describe the organization of these components and their roles in constructing the benchmark samples. Detailed instructions for downloading and using the data are available in the dataset repository.

\subsection{Final Benchmark Samples}

The processed benchmark samples are provided in two directories,  \\ \texttt{mono\_temporal\_samples} and \texttt{multi\_temporal\_samples}. Each directory contains five TAR archives corresponding to the training, validation, Test-Easy, Test-Medium, and Test-Hard subsets, as defined in Section~\ref{sec:dataset}. The directories also include separate JSON files containing the training-set normalization statistics for each configuration.

The final samples are generated through a two-step process. First, fire IDs are assigned to the training, validation, and test subsets using the fire growth points and metadata files, following the spatial and temporal splitting strategy described in the main text. Second, individual samples are constructed by iterating through the tile-specific keys in the metadata JSON files. For a given tile and day of burn, we apply the following rules:

\begin{itemize}
    \item \textbf{Mask merging:} If multiple fire events, or different portions of large fires, are active within the same tile, the pipeline accesses the corresponding HDF5 files, extracts their fire masks, and merges them into a unified mask representing the accumulated burned area within the $256 \times 256$ tile.

    \item \textbf{Feature extraction:} Since environmental observations for a given geographical tile and date are shared across overlapping fire events, the corresponding features are extracted from a single reference HDF5 file to avoid redundant data loading.
\end{itemize}

The mono-temporal configuration uses observations from a single day to predict the next-day wildfire spread, whereas the multi-temporal configuration incorporates observations from three consecutive days. Both configurations follow the same spatial and temporal data partitioning strategy. Normalization is applied using statistics computed exclusively from the corresponding training set, as described in Section~\ref{sec:dataset}.

To facilitate exploration without downloading the complete dataset, we also provide a representative sample and an accompanying Jupyter notebook in the \texttt{representative\_sample} directory.

\subsection{Raw Data Organization}

The raw dataset contains spatially harmonized but unnormalized data, allowing researchers to construct alternative input representations, investigate additional features, and develop new preprocessing strategies. It is organized into the following three components.

\subsubsection*{Fire Growth Points Data}

The fire growth points are provided as yearly CSV files obtained from the Canadian Fire Spread Dataset (CFSDS) \citep{CFSDS}. These files are distributed without modification and contain fire identifiers and geographical information, including the coordinates associated with individual fire events.

The fire identifiers and locations are used to associate events with the dataset partitions and support the spatial splitting procedure described in the main text.

\subsubsection*{Yearly Fire Metadata}

The yearly metadata are stored as JSON files that organize fire activity by geographical tile, year, and day of burn. Each entry is indexed by a composite key of the form \texttt{tileID\_year\_DOB}, where \texttt{tileID} identifies a fixed geographical tile and \texttt{DOB} denotes the day of year.

Each key maps to a list of fire identifiers and their corresponding day keys for events active within the specified tile and date.

This metadata structure serves two purposes. First, it supports the association of fire events and geographical tiles with their corresponding dataset partitions. Second, it enables the sample construction pipeline to identify overlapping fire events within the same tile and merge their fire masks into a unified representation.

\subsubsection*{HDF5 Fire Data}

The main geospatial data are stored in HDF5 (\texttt{.h5}) files, with one file per unique fire event, named according to its identifier (e.g., \texttt{fire\_2024\_188.h5}). Each file contains the following hierarchical structure:

\begin{itemize}
    \item \textbf{Global attributes:} Metadata stored at the file root, including the coordinate reference system, spatial resolution, fire identifier, and fire year.

    \item \textbf{Tile groups:} Each file contains separate groups corresponding to the geographical tiles intersected by the fire event. Each tile group includes:
    \begin{itemize}
        \item \textbf{\texttt{static\_features}:} Spatial environmental features that do not change daily, including topography, forest structure (SCANFI), land cover, disturbance history, and the Human Influence Index (HII).

        \item \textbf{\texttt{coords}:} Geographical coordinate grids for the tile, including latitude, longitude, easting, and northing.

        \item \textbf{\texttt{days}:} Day-specific observations organized by the corresponding fire day. Each day group contains:
        \begin{itemize}
            \item \textbf{\texttt{features}:} Daily environmental observations, including meteorological variables, vegetation indices, and the corresponding fire progression information.

            \item \textbf{\texttt{satellite}:} Sentinel-2 spectral bands and associated acquisition metadata, including observation dates and cloud coverage information.
        \end{itemize}
    \end{itemize}
\end{itemize}

The HDF5 files contain spatially harmonized but unnormalized observations. Environmental variables retain their values before benchmark normalization, and Sentinel-2 bands are stored without applying the $10^{-4}$ reflectance scaling factor. The original disturbance categories are also preserved in the raw release, while the processed benchmark samples use the leakage-controlled wildfire disturbance representation described in Section~\ref{sec:dataset}.

\subsection{Data Licensing and Attribution}

DailyWildfireCAN is distributed under the Creative Commons Attribution-NonCommercial-ShareAlike 4.0 International (CC BY-NC-SA 4.0) license for original dataset contributions and eligible adaptations. Third-party materials incorporated into the dataset remain subject to their respective original licenses.

Table~\ref{tab:licences} summarizes the licenses of the data sources used to construct DailyWildfireCAN. Source data have undergone preprocessing, including reprojection to EPSG:3347 and spatial resampling to $90$~m resolution. Additional preprocessing details are provided in Section~\ref{sec:dataset}.

\begin{table*}[htbp]
\centering
\caption{Original licenses of the data sources used to construct DailyWildfireCAN. Third-party materials remain subject to their respective licensing and attribution requirements.}
\label{tab:licences}

\small
\setlength{\tabcolsep}{5pt}
\renewcommand{\arraystretch}{1.1}

\begin{tabularx}{\linewidth}{@{}lX@{}}
\toprule
\textbf{Data Source} & \textbf{License} \\
\midrule

ASTER DEM \citep{aster_dem}
& \href{https://creativecommons.org/publicdomain/zero/1.0/}{CC0} \\
\addlinespace[4pt]

ERA5-Land \citep{munoz2021era5}
& \href{https://creativecommons.org/licenses/by/4.0/}{CC BY 4.0} \\
\addlinespace[4pt]

VIIRS \citep{VIIRS_VNP13A1_2021}
& \href{https://creativecommons.org/publicdomain/zero/1.0/}{CC0} \\
\addlinespace[4pt]

SCANFI \citep{scanfi}
& \href{https://open.canada.ca/en/open-government-licence-canada}{Open Government Licence -- Canada 2.0} \\
\addlinespace[4pt]

Land Cover \citep{landcover_ccrs}
& \href{https://open.canada.ca/en/open-government-licence-canada}{Open Government Licence -- Canada 2.0} \\
\addlinespace[4pt]

CanLaD \citep{canlad}
& \href{https://open.canada.ca/en/open-government-licence-canada}{Open Government Licence -- Canada 2.0} \\
\addlinespace[4pt]

Sentinel-2 \citep{drusch2012sentinel2}
& \href{https://sentinels.copernicus.eu/documents/247904/690755/Sentinel_Data_Legal_Notice}{Copernicus Sentinel Data Terms}. Contains modified Copernicus Sentinel data [2017--2024]. \\
\addlinespace[4pt]

HII \citep{sanderson2022march}
& \href{https://creativecommons.org/licenses/by-nc-sa/3.0/}{CC BY-NC-SA 3.0} \\
\addlinespace[4pt]

CFSDS \citep{CFSDS}
& \href{https://creativecommons.org/licenses/by/4.0/}{CC BY 4.0} \\

\bottomrule
\end{tabularx}
\end{table*}